\documentclass[journal]{IEEEtran}

\usepackage{ifpdf}

\usepackage{cite}
\usepackage{hyperref}
\hypersetup{
    colorlinks   = true,
    citecolor    = green
}
\ifCLASSINFOpdf
  \usepackage[pdftex]{graphicx}
  \graphicspath{{Figures/}}
  \DeclareGraphicsExtensions{.pdf,.jpeg,.png}
\else
  \DeclareGraphicsExtensions{.eps}
\fi

\DeclareGraphicsExtensions{.pdf,.eps,.jpeg,.png}
\usepackage{epstopdf}
\usepackage{epsfig}
\usepackage{subcaption}
\usepackage{float}
\usepackage{mathptmx}      
\usepackage{latexsym}

\usepackage{multirow}
\usepackage{amsmath}
\usepackage{amssymb}
\usepackage{mathrsfs}

\usepackage{url}
\usepackage[nolist]{acronym}
\usepackage{todonotes}

\usepackage{lscape}
\usepackage{algpseudocode}
\usepackage{algorithm}
\usepackage{array}

\usepackage{pdfpages}

\usepackage{pifont}
\newcommand{\cmark}{\ding{51}}%

\newcolumntype{L}[1]{>{\raggedright\let\newline\\\arraybackslash\hspace{0pt}}m{#1}}
\newcolumntype{C}[1]{>{\centering\let\newline\\\arraybackslash\hspace{0pt}}m{#1}}
\newcolumntype{R}[1]{>{\raggedleft\let\newline\\\arraybackslash\hspace{0pt}}m{#1}}

\begin{document}
\sloppy

\title{Robust Real-Time Multi-View Eye Tracking}

\author{Nuri~Murat~Arar,~\IEEEmembership{Student Member,~IEEE,}
        and~Jean-Philippe~Thiran,~\IEEEmembership{Senior Member,~IEEE}
\thanks{N. M. Arar and J.-P. Thiran are with the Signal Processing Laboratory (LTS5), \'Ecole Polytechnique F\'{e}d\'{e}rale de Lausanne (EPFL), Switzerland.}
}

\markboth{IEEE TRANSACTIONS ON CIRCUITS AND SYSTEMS FOR VIDEO TECHNOLOGY,~Vol.~XX, No.~XX, ~2018}%
{Shell \MakeLowercase{\textit{et al.}}: Bare Demo of IEEEtran.cls for IEEE Journals}

\maketitle

\acrodef{PoR}{point of regard}
\acrodef{fps}{frames-per-second}
\acrodef{NIR}{near-infrared}
\acrodef{FoV}{field-of-view}

\begin{abstract}
Despite significant advances in improving the gaze tracking accuracy under controlled conditions, the tracking robustness under real-world conditions, such as large head pose and movements, use of eyeglasses, illumination and eye type variations, remains a major challenge in eye tracking. In this paper, we revisit this challenge and introduce a real-time multi-camera eye tracking framework to improve the tracking robustness. First, differently from previous work, we design a multi-view tracking setup that allows for acquiring multiple eye appearances simultaneously. Leveraging multi-view appearances enables to more reliably detect gaze features under challenging conditions, particularly when they are obstructed in conventional single-view appearance due to large head movements or eyewear effects. The features extracted on various appearances are then used for estimating multiple gaze outputs. Second, we propose to combine estimated gaze outputs through an adaptive fusion mechanism to compute user's overall point of regard. The proposed mechanism firstly determines the estimation reliability of each gaze output according to user's momentary head pose and predicted gazing behavior, and then performs a reliability-based weighted fusion. We demonstrate the efficacy of our framework with extensive simulations and user experiments on a collected dataset featuring 20 subjects. Our results show that in comparison with state-of-the-art eye trackers, the proposed framework provides not only a significant enhancement in accuracy but also a notable robustness. Our prototype system runs at 30 frames-per-second (fps) and achieves $\sim$1$^\circ$ accuracy under challenging experimental scenarios, which makes it suitable for applications demanding high accuracy and robustness.

\end{abstract}

%
\IEEEpeerreviewmaketitle

\vspace{-6mm}
\section{Introduction}
\label{introduction}
Gaze movements provide cues indicating observer's visual attention, emotional state and cognitive processes\cite{Underwood2005,Duchowski2007}. Therefore, tracking gaze movements, also known as eye tracking, is essential for human behaviour research and diagnostics applied in a wide variety of disciplines, including among many others, sociology, psychology, cognitive science, neuroscience, and marketing research. Besides, it is an important modality to enhance human-computer interaction for controlling and navigation, such as in virtual reality, augmented reality, and gaming research. The popularity of eye tracking technology has recently been increasing, owing to its large spectrum of applications as well as the promising technical advancements. Despite valuable efforts, high-accuracy ($\leq$1$^\circ$) eye tracking systems still suffer from various factors, such as high cost, complex and inflexible setup configurations, and more importantly, low tolerance to varying real-world conditions, which hinder them from being widely used. Hence, there is room for further research efforts, particularly towards improving the tracking robustness under unconstrained conditions.

Remote video-oculography, in which users' eyes are non-intrusively captured by remote sensors, is the focus of this paper since it provides the most natural and convenient interaction for the users. As described in a recent survey \cite{Survey_Hansen2010}, remote sensor-based eye tracking methods can be classified into two categories, namely, \emph{appearance-based} and \emph{feature-based}. \emph{Appearance-based} methods use the image content as the input. They learn a mapping from the image features directly to the gaze points. On the other hand, \emph{feature-based} methods utilize local features extracted on eye images, such as pupil center and reflections on the cornea (aka glints), to determine the gaze. They mostly require particular hardware configuration and leverage the features that are formally related to the gaze points through the geometry of the system and eye physiology. \emph{Feature-based} methods can further be categorized into \emph{3D model-based}, \emph{regression-based}, and \emph{cross ratio-based} methods. Each category has its own advantages and disadvantages regarding the estimation accuracy, robustness, and system implementation complexity, as summarized in Table~\ref{table1}. \emph{3D model-based} methods \cite{Beymer2003,Hennessey2006,Guestrin2007,Park2007,Sun2015} compute the gaze from the features obtained from a 3D geometric eye model, whereas \emph{regression-based methods} \cite{Zhu2006,Zhu2007,Cerrolaza2008,Sesma-sanchez2012} assume a direct mapping from the features to the gaze points. On the other hand, \emph{cross ratio-based methods} \cite{Yoo2005,Hansen2010,Coutinho2013,Zhang2014,Huang2014,Arar2016j} compute the gaze by leveraging the cross ratio property of the projective space.


\begin{table*}[!t]
\small
\vspace{-3mm}
\caption{A generic comparison of gaze estimation techniques.}
\vspace{-2mm}
\label{table1}
\centering
\begin{tabular}{l|ccc|c}
\noalign{\smallskip}
Evaluation Criteria	& 3D Model-based & Regression-based & Cross Ratio-based & Appearance-based  \\ 
\noalign{\smallskip}\hline\noalign{\smallskip}
Setup complexity:	        & high & medium & medium & low \\ 
System calibration:	 		& fully-calibrated & $\times$ & $\times$ & $\times$ \\ 
Hardware requirements (\# of cameras): & 2+ infrared (stereo) & 1+ infrared & 1+ infrared & 1+ ordinary \\
Hardware requirements (\# of lights):  & 2+ infrared & 2+ infrared & 4+ infrared & $\times$ \\
Gaze estimation accuracy error: & $<1^\circ$  & $\sim1-2^\circ$ & $\sim1-2^\circ$ & $>2^\circ$ \\ 
Implicit robustness to head movements: & medium-high & low-medium & low-medium & low \\ 
Implicit robustness to varying illumination: & medium-high  & medium-high  & medium-high & low  \\ 
Implicit robustness to use of eyewear: & low & low  & low & medium  \\ 
\noalign{\smallskip}\hline
\end{tabular}
\vspace{-3mm}
\end{table*}

Over more than two decades, numerous works from each of the aforementioned categories have been presented. Among these, the main emphasis has been mostly given to the estimation accuracy improvements through introducing different gaze models \cite{Yoo2005,Guestrin2007,Zhu2007,Arar2015f} or developing effective user calibration techniques \cite{Kang2007,Villanueva2008,Huang2014,Arar2015w,Arar2016j}. Consequently, high accuracies ($<$1$^\circ$) are reported under controlled conditions. Nevertheless, research on the tracking robustness under real-world conditions, such as large head movements, use of eyewear, variations in illumination and eye type, have been largely neglected. Thus, these remain major concerns in eye tracking. 

In this paper, extending our previous efforts that focus on improving the estimation accuracy \cite{Arar2015f,Arar2016j}, we address the major robustness concerns of existing eye tracking systems. We present a real-time multi-view gaze estimation methodology to improve the tracking robustness to real-world conditions. Instead of tracking the gaze from a single view as performed by previous work, we design a multi-view framework to leverage multiple eye appearances simultaneously acquired from various views. In conventional single-view setups, there exists a single appearance, on which the features may be obstructed due to challenging conditions such as large head movements or occlusions caused by glasses. Whereas in our framework, the main benefit is to simultaneously perform feature detection on multi-view appearances. For each frame, our approach enables to compute multiple gaze outputs using the extracted features. Furthermore, these gaze outputs are effectively combined by a novel adaptive fusion mechanism to compute user's overall \ac{PoR}. In this context, the proposed mechanism firstly determines the estimation reliability of each gaze output according to certain gaze reliability indicators, e.g., user's predicted gazing behavior, momentary head pose with respect to the camera. Then, it performs a reliability-based weighted fusion, which leads to improved accuracy and robustness. Extensive evaluations on both simulated and real-world data were performed to validate the proposed methodology. In addition to thorough simulations, a database featuring 20 users performing 8 experiments under varying illumination conditions and head movements was collected. In these experiments, natural human-computer interaction was targeted. Users followed some conventional experimental scenarios as well as a newly introduced one. The results demonstrate that in comparison to conventional single-view eye tracking, the multi-view approach provides a significantly better performance both in accuracy and robustness to aforementioned challenging conditions. 

The proposed framework is highly flexible and can easily adapt to hardware and software modifications. Depending on the application type and desired tracking performance, the employed gaze estimation method, number of cameras and their configuration can simply be alternated, even without requiring any system adjustments (e.g., camera or geometric scene calibration). Our current uncalibrated prototype relies on a \emph{cross ratio-based} gaze estimation method, and operates with low-resolution eye data. The system's overall hardware setup and computational complexities are lower than those of fully-calibrated systems. Hence, it enables a fast and accurate eye tracking without requiring any cumbersome camera and geometric scene calibrations. Our three-camera prototype outputs \ac{PoR}s with an accuracy of $\sim$1$^\circ$ at 30 fps and also obtains nearly 100\% estimation availability under challenging scenarios.

The outline of the paper is described as follows: Section~\ref{related_work}
gives an overview of the related work. Section~\ref{proposed_system} describes the proposed framework. Evaluations on the simulated and real data are presented in Section~\ref{evaluation_simulation} and \ref{evaluation_real}. Section~\ref{discussion} discusses the acquired insights, and conclusions are given in Section~\ref{conclusion}.

\vspace{-3mm}
\section{Related Work}
\label{related_work}


A generic comparison of existing eye tracking solutions regarding various evaluation criteria is given in Table~\ref{table1}. The following presents a detailed overview of the related work. 


\vspace{-3mm}
\subsection{Gaze Estimation Accuracy \& Setup Complexity}
The majority of the existing work focus on improving the estimation accuracy. There is no doubt that the accuracy is directly proportional with the setup complexity. \emph{3D model-based} methods \cite{Beymer2003,Guestrin2007,Park2007,Hennessey2006,Lai2015} are widely preferred as they provide high accuracy under generic head movements, owing to their explicit and fine 3D modeling. Most commercial high-accuracy solutions rely on \emph{3D model-based} methods. However, they have a significant drawback, that is to require fully-calibrated systems. To acquire an accurate 3D eye model, a complex setup that requires camera and geometric scene calibrations (e.g., stereo, depth sensor) is needed. Alternatively, \emph{cross ratio-based} \cite{Yoo2005,Hansen2010,Coutinho2013,Zhang2014,Huang2014,Arar2016j,Arar2015f} and \emph{regression-based} methods \cite{Cerrolaza2012,Sesma-sanchez2012} have mostly lower setup complexity and avoid setup calibrations. However, they rely on approximations, and so, their performances are lower in accuracy and movement robustness. On the contrary to \emph{feature-based} methods, \emph{appearance-based} methods \cite{Lu2015,Zhang2015,Krafka2016,Wood2016a} simply require an ordinary camera. Yet, they are restricted to particular applications due to their limitations in accuracy and robustness.

The great majority of the existing eye trackers, regardless of the gaze estimation method employed, rely on a single-view framework, which employs either a single-camera setup \cite{Zhang2015,Sesma-sanchez2012,Hansen2010,Coutinho2013}, or a multi-camera setup that is designed to acquire 3D info (stereo, depth) \cite{Guestrin2006,Hennessey2006,Zhu2007,Lai2015} or to acquire high-resolution eye data using a pan-tilt unit \cite{Beymer2003,Park2007,Yoo2005}. On the other hand, the efficacy of the multi-camera systems that perform multi-view tracking has not adequately been investigated. In this regard, to the best of our knowledge, there exists only one previous effort. \cite{Utsumi2012} proposed a two-camera setup mainly to obtain a wide observation area for a gaze-reactive signboard. Yet, the system was designed for a highly coarse gaze tracking, which achieved $\sim$11$^\circ$ accuracy. Hence, we are the first to exploit a multi-camera setup for multi-view tracking so as to improve the estimation performance for high-accuracy eye tracking \cite{Arar2015f}. In this paper, we extend our previous work to also improve the tracking robustness under real-world conditions by investigating various multi-camera configurations as well as novel adaptive fusion mechanisms.

\vspace{-3mm}
\subsection{User Calibration}
In addition to hardware setup calibration, user calibration plays an important role in user experience and convenience. User calibration is required for modelling the person-specific eye parameters, which are crucial for the estimation bias correction. The calibration quality improves, to a certain extent, when the amount of calibration data increases. However, augmenting the data amount by increasing the number of calibration points could be tedious and harms the user experience. In this regard, the trade-off between the quality and convenience of user calibration has been widely studied in the literature. Significant advancements have been made, for instance, better geometric eye models \cite{Guestrin2006,Villanueva2008,Nagamatsu2011}, more effective bias correction models \cite{Hansen2010,Zhang2014,Arar2015w,Arar2016j}, and implicit calibration methods \cite{Sun2014,Chen2015} were developed. 

\subsection{Head Movement Robustness}
\vspace{-1mm}
\emph{3D model-based}  methods are theoretically more tolerant to the changes in head pose and location due to explicit parametrization of person-specific eye parameters. Yet, in practice, they suffer from inaccuracy under large head movements. One of the main reasons is that most systems are faced with the dilemma of trading off between the head movement range and data resolution. In early efforts \cite{Beymer2003,Park2007}, a wide \ac{FoV} stereo system was employed to allow free head movement together with one or more narrow \ac{FoV} stereo system to capture high-resolution eye images. These systems were mostly interconnected through a pan-tilt unit which mechanically reoriented the narrow \ac{FoV} camera to users' eye. Despite enabling high accuracy and certain head movement robustness, the use of a pan-tilt unit increased the setup complexity and cost. Later, researchers avoided such mechanical units and focused on introducing more robust models, which eliminated the need for narrow \ac{FoV} cameras. For instance, \cite{Guestrin2007} introduced a method that used the pupil center and at least two glints, which were estimated from the eye images captured by at least two cameras. Their system achieved $<$1$^\circ$ accuracy by tolerating head movements in a volume of 10$\times$8$\times$10 cm$^3$. In addition, \cite{Hennessey2006} presented a single camera non-stereo system that employed ray tracing rather than depth from focus. Their system allowed an accurate ($<$1$^\circ$) estimation in a volume of 14$\times$12$\times$20 cm$^3$. Recently, \cite{Sun2015} proposed a Kinect sensor-based technique, which used a parametrized iris model. They reported 1.4$-$2.7$^\circ$ accuracy error under head movements in a volume of 20$\times$20$\times$8 cm$^3$.

\emph{Regression-based} methods indirectly model the eye physiology, geometry, and optical properties. When the user moves away from the calibration position, the features non-linearly change, therefore, the calibration mapping becomes less accurate causing inaccurate estimations. To address this, multiple glints-based approaches have been suggested. \cite{White1993} proposed to use a second light source, which permitted differentiation of head movement from eye rotation in the camera image. Using two glints as points of reference and exploiting spatial symmetries, they proposed a spatially dynamic calibration method to compensate for lateral head translation automatically. Later, a thorough review of polynomial-based regression methods using two glints was presented in \cite{Cerrolaza2008}. They evaluated various models using different pupil-glint vectors and polynomial functions. In addition, \cite{Sesma-sanchez2012} studied how binocular information can improve the accuracy and robustness against head movements for the polynomial based systems using one or two glints. Moreover, \cite{Cerrolaza2012} suggested two calibration strategies to reduce the errors caused by head movements. The results of the experiments showed that both strategies achieved a reduction in error by a factor of two for $\pm$6 cm depth movements. From a different perspective, \cite{Zhu2007} proposed a stereo-based system that achieved an $\sim$2$^\circ$ accuracy while allowing for a significantly greater working volume (20$\times$20$\times$30 cm$^3$) without using a chinrest. They estimated 3D optical eye axis by directly applying triangulation techniques on the glints and pupil center. They also suggested that 3D head pose can be used to compensate for the bias caused by head movements. However, the main drawback was that a multi-camera fully-calibrated stereo setup was required to obtain 3D information.

\emph{Cross ratio-based} methods are sensitive mainly to the depth movements. Various attempts have been made to enhance the depth movement tolerance. Most of these focused on adapting the user calibration to the changes in head movements. For instance, \cite{Coutinho2013} proposed dynamic calibration correction and planarization of features, which achieved $\sim$0.5$^\circ$ accuracy while tolerating up to $\pm$12.5 cm depth changes. However, their system required high-resolution (640$\times$480 pixels) eye images. Also, a chinrest was required to keep users' eye within camera's \ac{FoV} and to fix users' head pose and location during the experiments. \cite{Zhang2014} proposed a homography-based calibration modeling with a binocular fixation constraint to jointly estimate the homography matrix from both eyes. They reported $\sim$0.6$^\circ$ using a much lower resolution while allowing for $\pm$5 cm head movements. A potential drawback of their system was that the features from both eyes must be accurately detected to compute the gaze, which constrains the estimation availability due to the limited head pose allowance. Moreover, \cite{Huang2014} proposed an adaptive homography calibration. They learned an offline model on the simulated data by exploring the relationship between the estimation bias and varying head movements. They achieved promising results on the simulated ($\pm25$ cm) and real data ($\pm$10 cm). An important limitation in \cite{Zhang2014} and \cite{Huang2014} is that they use a chinrest to fix the head pose during the evaluations, similar to \cite{Coutinho2013}. Although reporting performances with chinrest leads to more stable results, it causes the evaluations to discard the impact of continuous head pose variations. Besides, it is impractical for real-world applications and significantly harms user experience. 

\begin{figure*}[t!]
\centering
\vspace{-3mm}
\includegraphics[width=2\columnwidth]{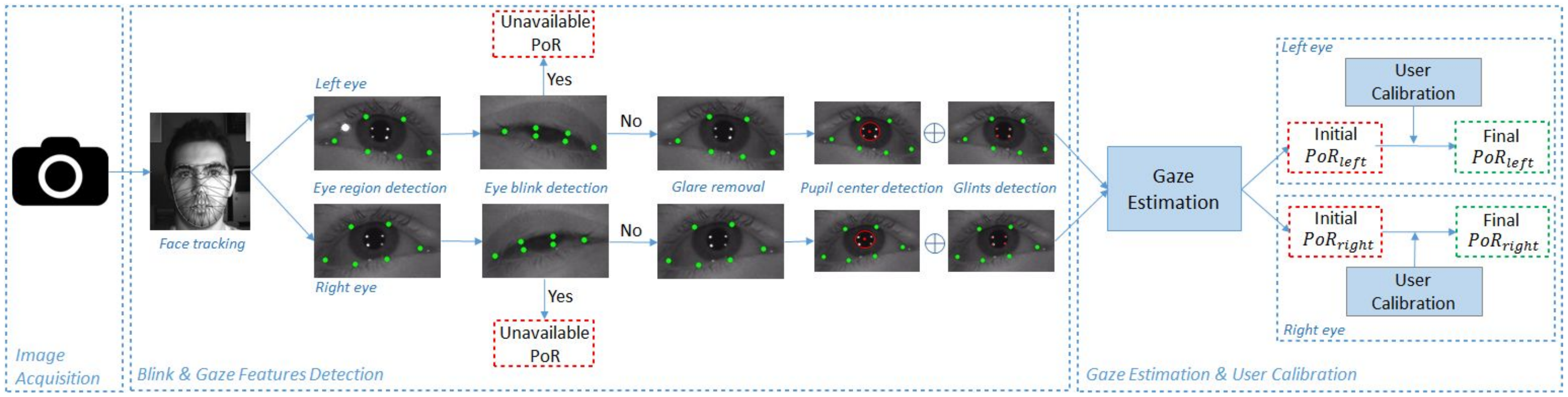}
\vspace{-2mm}
\caption{Overview of a single-camera system, which comprises of two gaze sensors, one for each eye.}
\label{single_camera_system} 
\vspace{-6mm}
\end{figure*}

\vspace{-4mm}
\subsection{Eyewear (Glasses) Robustness}
\vspace{-1mm}
\label{related_work_eye_wear}
Eyewear robustness, particularly to glasses, has been a challenging research problem since the reflection and refraction from glasses drastically obstruct the eye appearance and features. As \emph{appearance-based methods} \cite{Zhang2015,Wood2016a,Krafka2016} neither use light sources nor rely directly on the detection of individual gaze features, their performances are less affected by the glasses in comparison with \emph{feature-based methods}, which require explicit solutions. Unfortunately, glasses robustness has largely been neglected by the majority of the previous work. There exists only a limited number of attempts to address the reflections and refraction on the glasses. \cite{Ebisawa1998} introduced a robust pupil detection method by leveraging the bright-pupil effect generated with a differential lighting scheme. He also suggested a method for eliminating the reflections appearing on the glasses. His method was successfully realized in \cite{Ji2002} for monitoring driver vigilance. In addition, \cite{Park2007} proposed a dual illumination technique to avoid the reflections on the glasses. When a specular reflection was detected, the system deactivated the current illuminator and activated the alternative illuminator on the opposite side, such that the reflection can be avoided. Furthermore, \cite{Guestrin2006,Villanueva2007} demonstrated that compensating for the refraction can increase the accuracy up to $\sim$1$^\circ$. Recently, \cite{Kubler2016} also simulated reflection and refraction of glasses to study their impacts in connection with pupil and glint detection. In our work, from a different perspective, we address the robustness to glasses by focusing on generating and detecting more reliable gaze features. As the occlusion and distortion of features on an eye appearance depends on the relative positioning of a camera, light sources, and eye, we propose to perform multi-view tracking to obtain alternative eye appearances. Hence, the main benefit is that in case the glasses effects obstruct the features from certain views, they can still be recovered from alternative views.


\vspace{-3mm}
\section{Proposed Methodology}
\label{proposed_system}
\begin{figure}[t!]
\centering
\includegraphics[width=\columnwidth]{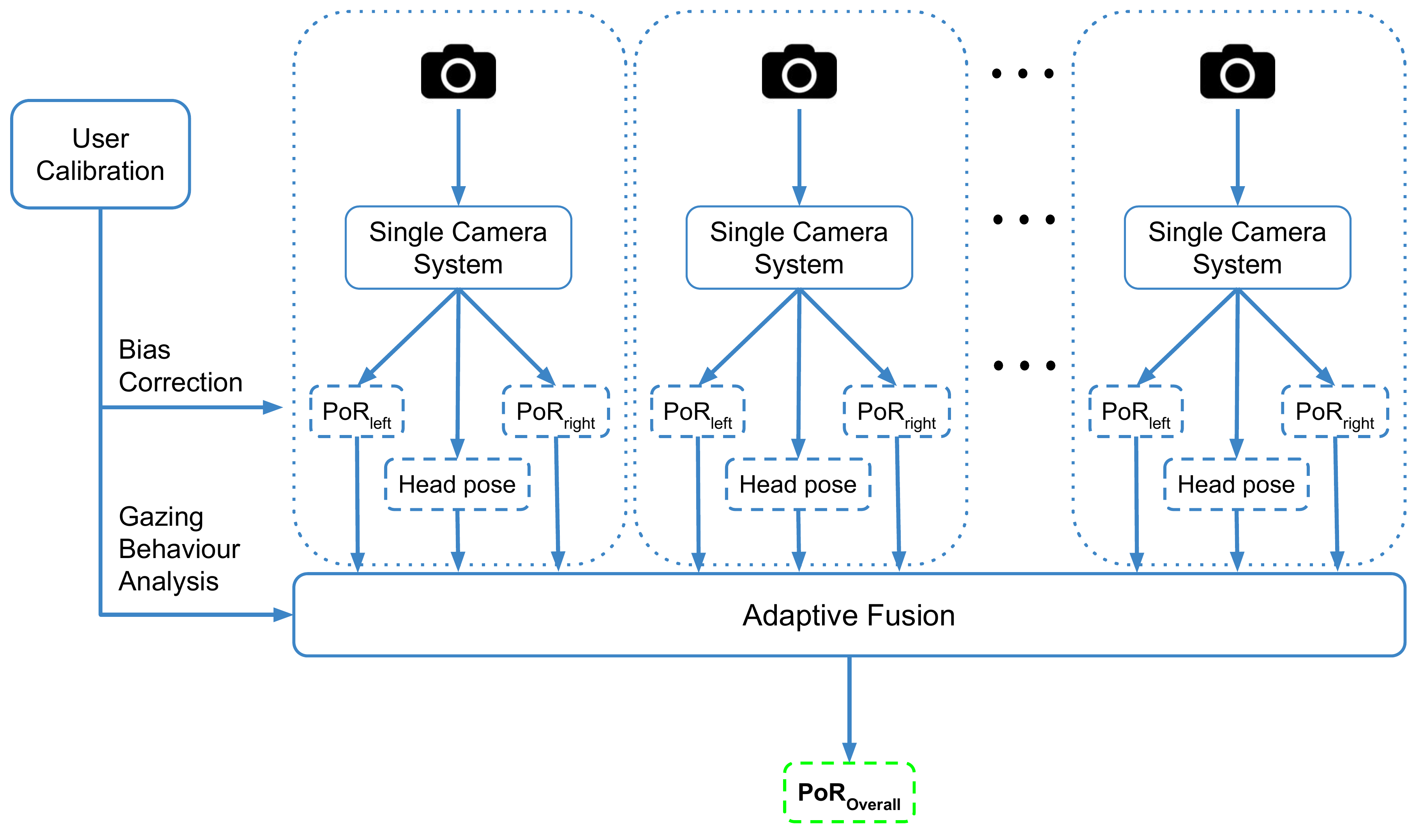}
\vspace{-6mm}
\caption{Overview of the proposed multi-camera framework.}
\label{multi_camera_system}
\vspace{-3mm}
\end{figure}

Our methodology comprises of simultaneously operating independent single-camera systems as illustrated in Fig.~\ref{multi_camera_system}. Each camera system firstly performs blink \& gaze features detection (Section~\ref{gaze_features_detection}) followed by gaze estimation and user calibration (Section~\ref{gaze_estimation}), as shown in Fig.~\ref{single_camera_system}. As the main contribution, gaze outputs obtained from each eye view from each camera are then fed into the proposed adaptive fusion mechanism to output the overall \ac{PoR} (Section~\ref{adaptive_fusion}). 


\vspace{-4mm}
\subsection{Blink \& Gaze Features Detection}
\label{gaze_features_detection}
Our methodology starts with eye localization, in which the existence of eyes is determined. We employ a robust non-rigid face tracker based on supervised decent method (SDM) \cite{sdm} to localize and track the eyes. SDM method assumes that an accurate final face shape with 66 landmarks can be estimated with a cascade of regression models given an initial shape. Once the shape is fitted accurately, we extract eye regions by using the landmarks around eyes. Note that neither registration nor alignment is required, i.e., no particular eye resolution is ensured. We detect eye blinks by computing the ratio of vertical eye opening to the eye width using the landmarks. If there is no blink, we continue with detection and removal of the specular reflections (glares) caused by glasses using well-known image processing techniques. Next, gaze features, i.e., corneal reflections (glints) and pupil center, are localized.

For glint detection, we initially perform histogram equalization to improve the contrast on input eye region. We then apply a thresholding to segment out the prospective glints. Here, we utilize spatial adaptive thresholding to take into account spatial variations in illumination. Instead of tuning a global threshold value, spatial adaptive thresholding applies different thresholds for small regions within the image, which leads to more robust results under varying illumination. Morphological opening and closing operations are then performed to get rid of the small blobs caused by noise. In the resulting binary image, we expect to find four blobs forming a trapezium since they emerge from the reflections of four LEDs located around the screen. Therefore, we perform connected component analysis to determine the candidate glints. If there are four or more candidate glints, we consider the shapes formed by any four-glints combination. The set of candidates whose convex hull has the highest match with a template shape representing the screen is considered as the final set of glints.


For pupil center detection, we follow a dark-pupil based approach rather than a bright-pupil based one due to its improved robustness to glasses and the variations in illumination and eye type \cite{ArarThesis2017}. More specifically, we first perform bilateral filtering on input eye region with dark-pupil to smooth the pupil region while still keeping the edges (pupil to iris) sharp. We then equalize the histogram to enhance the contrast. We approximate the average intensity within the pupil by the surrounding regions of each glint and center of glints polygon. We then remove the glints by filling them with the average intensity. On the resulting image, we apply global thresholding by considering the average intensity within the pupil. We then invert the image to highlight the pupil blob. Nonetheless, a few other blobs, which are as dark as the pupil region, such as eye lashes, eye lids, shades, also remain in the binary image. To distinguish the actual pupil region from the noisy blobs, we perform morphological operations for the noise removal. Among the remaining candidate blobs, we determine the final pupil by considering the shape, size, and location of the blobs. Its center of gravity is then used as the pupil center feature. Further explanations and figures on aforementioned processes can be found in \cite{ArarThesis2017}.


\vspace{-4mm}
\subsection{Gaze Estimation \& User Calibration}
\label{gaze_estimation}
It is important to note that the proposed multi-view framework is independent of the gaze estimation algorithm used. Therefore, any gaze estimation technique described in Section~\ref{related_work} can be utilized within this framework. The optimal method can be determined based on the application requirements, e.g., accuracy, robustness, setup complexity and flexibility. For instance, in our prototype system where the focus is high-accuracy \ac{PoR} estimation on desktop scenarios, we employ a \emph{cross ratio-based} method due to its particular advantages, such as enabling high-accuracy using an uncalibrated and flexible setup.

The original method relies on the cross-ratio invariant of the projective space \cite{Yoo2005}. More specifically, four light sources are positioned around a screen to create glints on subject's cornea. The polygon formed by the glints on the cornea is the projection of the screen. Another projection takes place from the corneal plane to the image plane. As the virtual tangent plane on the cornea has the same planar projective transformation of the screen and image planes, the pupil center on the image plane corresponds to the \ac{PoR} on the screen plane. Hence, the \ac{PoR} can be computed by the equality of the cross-ratios on screen and camera image planes, as detailed in \cite{ArarThesis2017}. Nevertheless, a user calibration is essential to compensate for the estimation bias caused by person-specific eye parameters, such as the angular offset between the visual and optical axis of the eyeball and the cornea radius and curvature. 
The calibration is performed once, prior to the use of the system. The users are asked to look at $N$ calibration points on the screen for $K$ frames long. Person-specific bias correction, $\mathcal{F}$, can be learned by minimizing the distances between the estimated gaze positions and corresponding calibration points on the screen as follows:
\begin{equation}
min \sum_{i}^{N} \sum_{j}^{K} \| {\bf P}_{i,j}  - \mathcal{F}({\bf Z}_{i,j}) \|,
\label{calibration_eq}
\end{equation}
where ${\bf Z}_{i,j}$ and ${\bf P}_{i,j}$ are the estimated \ac{PoR}s on the screen and the corresponding target calibration points, respectively. 

\begin{figure*}[t!]
\centering
\vspace{-3mm}
\includegraphics[width=1.6\columnwidth]{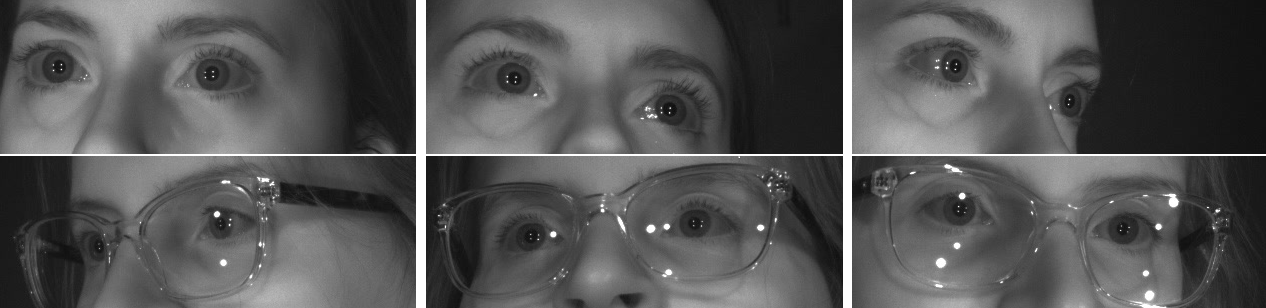}
\vspace{-1mm}
\caption{Example eye appearances from three different camera views: (left) left side camera, (middle) bottom camera and (right) right side camera while gazing at a stimulus point displayed on (top) upper-left and (bottom) upper-right region of the screen.}
\label{fig_murat-marili-chr}
\vspace{-3mm}
\end{figure*}

In this work, we apply a regularized least-squares regression based calibration technique \cite{Arar2016j}. The method has recently been shown to have better modeling and generalization capabilities than the state-of-the-art methods, owing to reduced model parameters. It enables to model the bias more effectively, particularly when the calibration data is limited in size and quality. Therefore, it facilitates tracking with low-resolution data and requires minimal user effort.

\vspace{-3mm}
\subsection{Adaptive Fusion Scheme}
\label{adaptive_fusion}
The proposed multi-view framework, which consists of individual single-camera trackers, is designed to empower a robust tracking under challenging conditions. Within this framework, each tracker simultaneously estimates the gaze for each eye (aka gaze sensor). In each frame, two distinct gaze outputs can be computed per camera. Consequently, in a multi-camera setup with C cameras, up to 2C gaze outputs can be generated per frame. The overall \ac{PoR} can then be computed by the fusion of available gaze outputs obtained from all sensors. Here, we propose an adaptive fusion mechanism to effectively combine the gaze outputs towards achieving a higher overall estimation accuracy and robustness. 

We have investigated several algorithms to perform the fusion. Among these, the most straightforward one is to average all gaze outputs. Despite its simplicity, fusion by simple averaging results in a significant improvement in estimation accuracy in comparison to the best of single-view trackers, particularly when the majority of the sensors produce reliable gaze outputs. In such cases, simple averaging provides a more accurate and consistent overall estimation through smoothing out the arbitrary noise. On the other hand, under more challenging scenarios, in which a higher variance exists among available gaze outputs, fusion by simple averaging is far from the optimal fusion. To this effect, we propose to combine the available gaze outputs in a weighted manner, where the weights are determined according to the estimation reliability of each sensor. Here, there is several factors that affect the reliability, such as camera's viewing angle, gaze point location on the screen, eyewear effects, person-specific gaze behaviours. Fig.~\ref{fig_murat-marili-chr} shows sample eye appearances captured from different camera views when users gaze at different regions on the screen.  While some views permit reliable gaze features, some others do not even contain any available features. The views capturing the best eye appearances continuously vary when users gaze at different regions on the screen. In addition, when users wear glasses, reflections and refraction occur on the glasses. When these effects distort or overlap with the features, the estimation becomes unavailable from that particular view. On the other hand, as there exists simultaneously captured several other views, the features can still be recovered from some of these views. This, in fact, constitutes an important benefit of the multi-view approach in comparison to conventional single-view approach, employed by the majority of the previous work. Hence, an effective fusion that accounts for the estimation reliability of individual gaze outputs can significantly improve the overall estimation accuracy and robustness. 

The adaptive fusion is formulated as follows:
\vspace{-1mm}
\begin{eqnarray}
\label{adaptivefusion}
{\bf z^{*}} &=& \sum_c \sum_e {\bf z_c^e}w_c^e, \\
\sum_c \sum_e w_c^e &=&1, \hspace{2mm}  e \in\{L,R\},~\hspace{2mm} c \in\{1,2,..,C\}, \nonumber
\end{eqnarray}
where ${\bf z^{*}}$ is the overall PoR and, $w_c^R$ and $w_c^L$ are the weights for the right and left eye's gaze outputs from the $c^{th}$ camera, respectively. In case any of the gaze outputs cannot be computed, then the weight of the missing one is set to zero. We do not report an overall \ac{PoR} in case there is no available gaze outputs for a given frame. To determine the weights, we propose two algorithms, namely, head pose-based fusion and person-specific gazing behaviour-based fusion: 





\subsubsection{Head pose-based fusion}
As can be depicted in Fig.~\ref{fig_murat-marili-chr} and psycho-visually evidenced in \cite{Zangemeister1982}, prior to the fixation, most users initially perform head rotation to find the most comfortable viewing angle. Therefore, when users gaze at different target points, the eye appearances continuously vary. Here, the estimation accuracy is strongly correlated with the quality of the eye appearance and gaze features, which relies on user's head pose relative to each camera view. For instance, when users gaze at the upper left corner of the screen, the left camera system often generates more accurate estimations than the others. The main reason is obviously that the relative head pose is more frontal from left camera's view, and consequently the corresponding eye appearances facilitate a more reliable feature detection. Hence, our first adaptive fusion relies on relative head pose angles estimated with respect to each camera. We assign the weights inversely proportional to the angles, as follows:
\begin{eqnarray}
\lambda_c^e &=& \frac{\alpha_{max}-|\alpha_c^e|}{\alpha_{max}}, \label{headposeweighting} \\
w_c^e &=& \frac{\lambda_c^e}{\sum_c \sum_e \lambda_c^e}, \label{normalization}
\end{eqnarray}
where $\alpha_c$ is the head pose yaw angle, $\alpha_{max}$ is the maximum angle allowed, e.g., 45$^\circ$. We calculate head pose angles using the landmarks obtained by the face tracker and point distribution model \cite{Saragih2009}. Both eyes are assigned with the same weights. Normalization is then performed using Eq.~(\ref{normalization}) prior to fusion.

Note also that we investigated an alternative weighting approach in our previous effort \cite{Arar2015f}. Instead of calculating head pose angles, we first calculate an initial \ac{PoR} using simple averaging. We then iteratively refine the initial estimation by weighting the cameras with respect to their distances to the estimations. Although this approach performs an effective fusion, it has two drawbacks: first, it requires camera locations to be known to compute the distances, and second, its performance is affected by the quality of the initial \ac{PoR}. A poor initial \ac{PoR} estimation leads to a less effective weighting. 

\subsubsection{Gazing behaviour-based fusion}
Although head pose-based weighting works well for most users, it does not take person-specific gazing behaviors into account, and consequently, it may experience a performance drop when a user has a particular gazing behaviour. For instance, although the majority of the users perform head rotation prior to fixation to have a comfortable viewing angle (frontal eyeball pose), some users do not perform any head movements but rather rotate their eye balls (non-frontal eyeball pose). In addition, head pose-based approach weighs cameras rather than the eyes. However, some users' vision may rely more on one particular eye than the other due to eye dominance or a physiological reason (e.g., lazy eye, strabism). For such users, assigning equal weights to both eyes may result in a low estimation performance. Hence, we propose to determine person-specific weights for each eye independently through leveraging user's calibration data. During user calibration, we generate fusion weight maps in addition to learning the user calibration model. Once the weight maps are obtained, our algorithm performs a weighted averaging of individual gaze outputs as follows:
\begin{eqnarray}
{\bf z^{*}} &=& \sum_c \sum_e {\bf z_c^e} \, {\bf M_c^e}({\bf z_c^e}.x,{\bf z_c^e}.y), \label{weightcomputation}\\
\sum_c \sum_e {\bf M_c^e}(x,y) &=&1, \hspace{2mm}  e \in\{L,R\},~\hspace{2mm} c \in\{1,2,..,C\}, \nonumber
\end{eqnarray}
where ${\bf z^{*}}$ is the overall \ac{PoR}, ${\bf z_c^e}$ are initial gaze outputs estimated using simple averaging, and ${\bf M_c^R}$ and ${\bf M_c^L}$ are the weight maps of right and left eye of the $c^{th}$ camera, respectively.

\begin{figure*}[t!]
\centering
\vspace{-3mm}
\subcaptionbox{$\textbf{M}_{\textbf{left~cam}}^{\textbf{L}}$ \label{fig_cal5}}{\includegraphics[width=0.33\columnwidth]{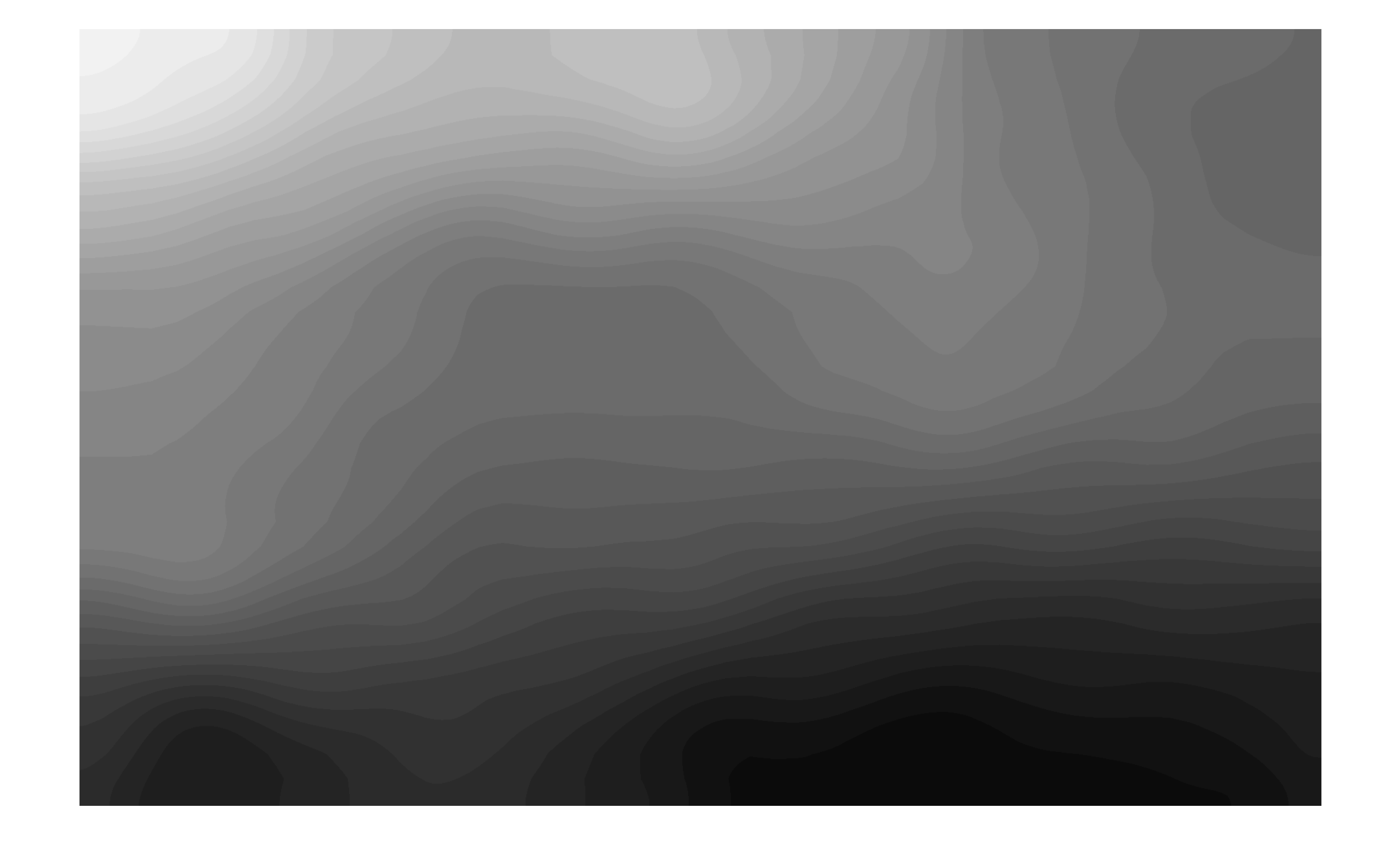}}
\subcaptionbox{$\textbf{M}_{\textbf{left~cam}}^{\textbf{R}}$ \label{fig_cal4}}{\includegraphics[width=0.33\columnwidth]{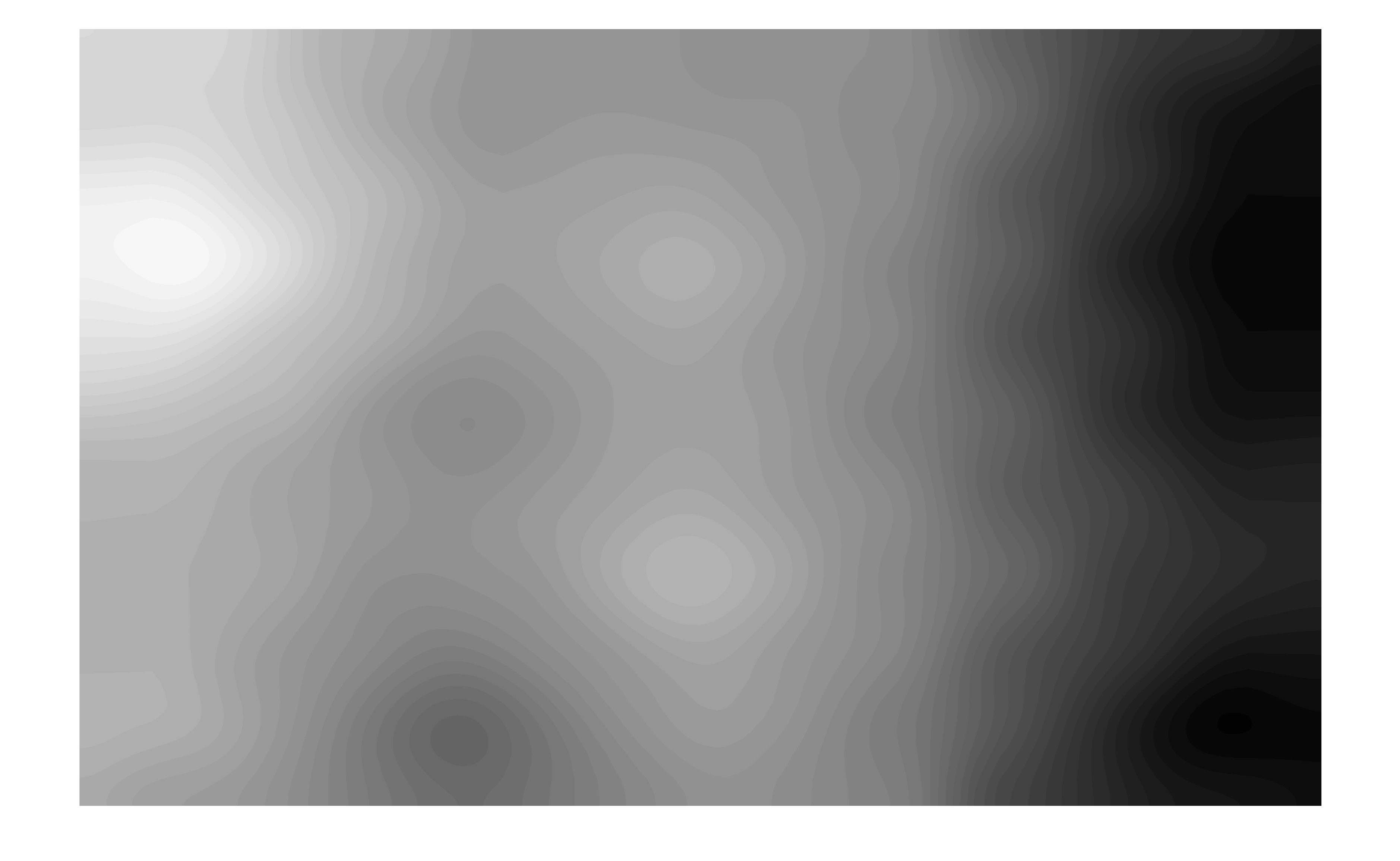}}
\subcaptionbox{$\textbf{M}_{\textbf{bottom~cam}}^{\textbf{L}}$ \label{fig_cal1}}{\includegraphics[width=0.33\columnwidth]{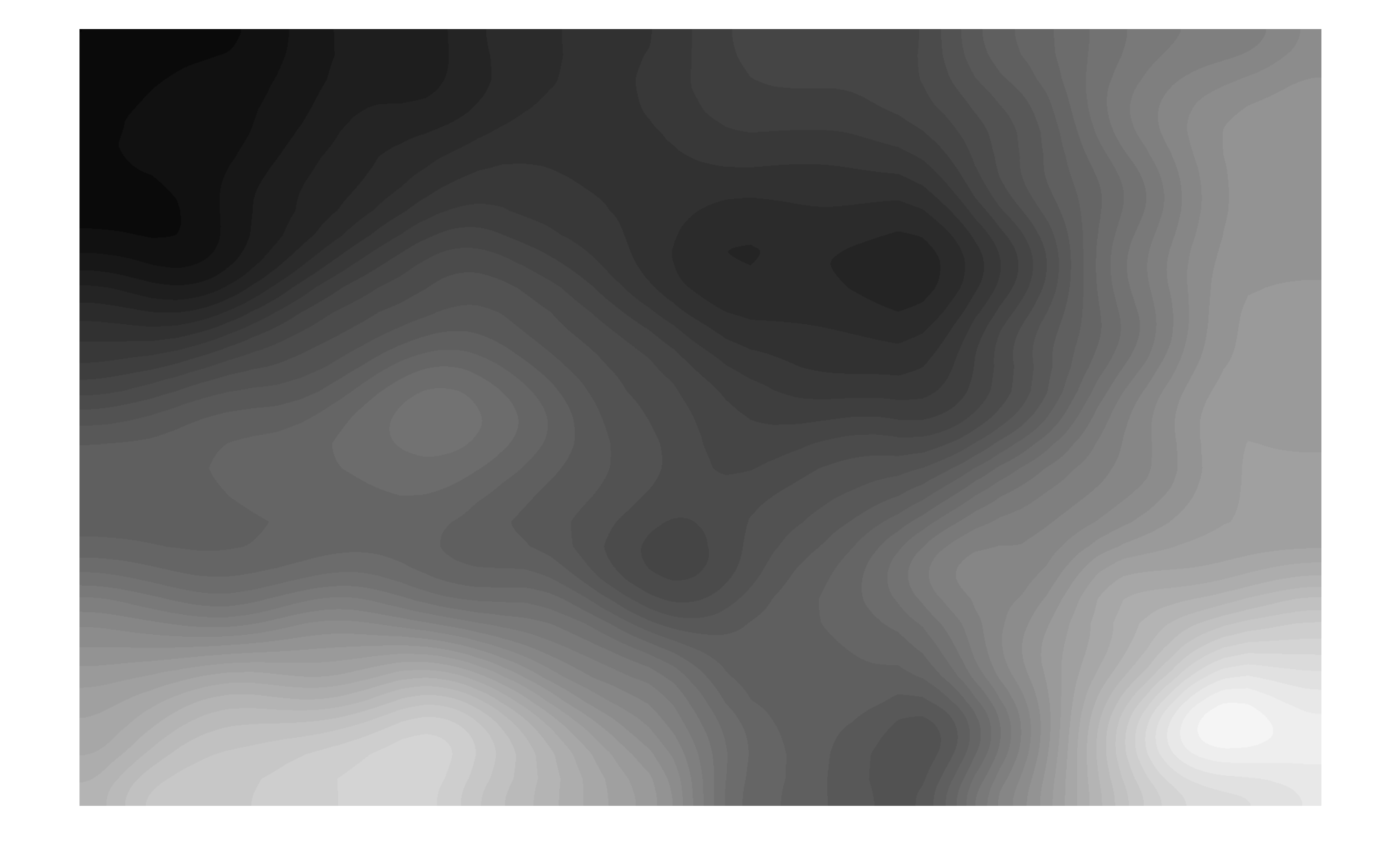}}
\subcaptionbox{$\textbf{M}_{\textbf{bottom~cam}}^{\textbf{R}}$ \label{fig_cal0}}{\includegraphics[width=0.33\columnwidth]{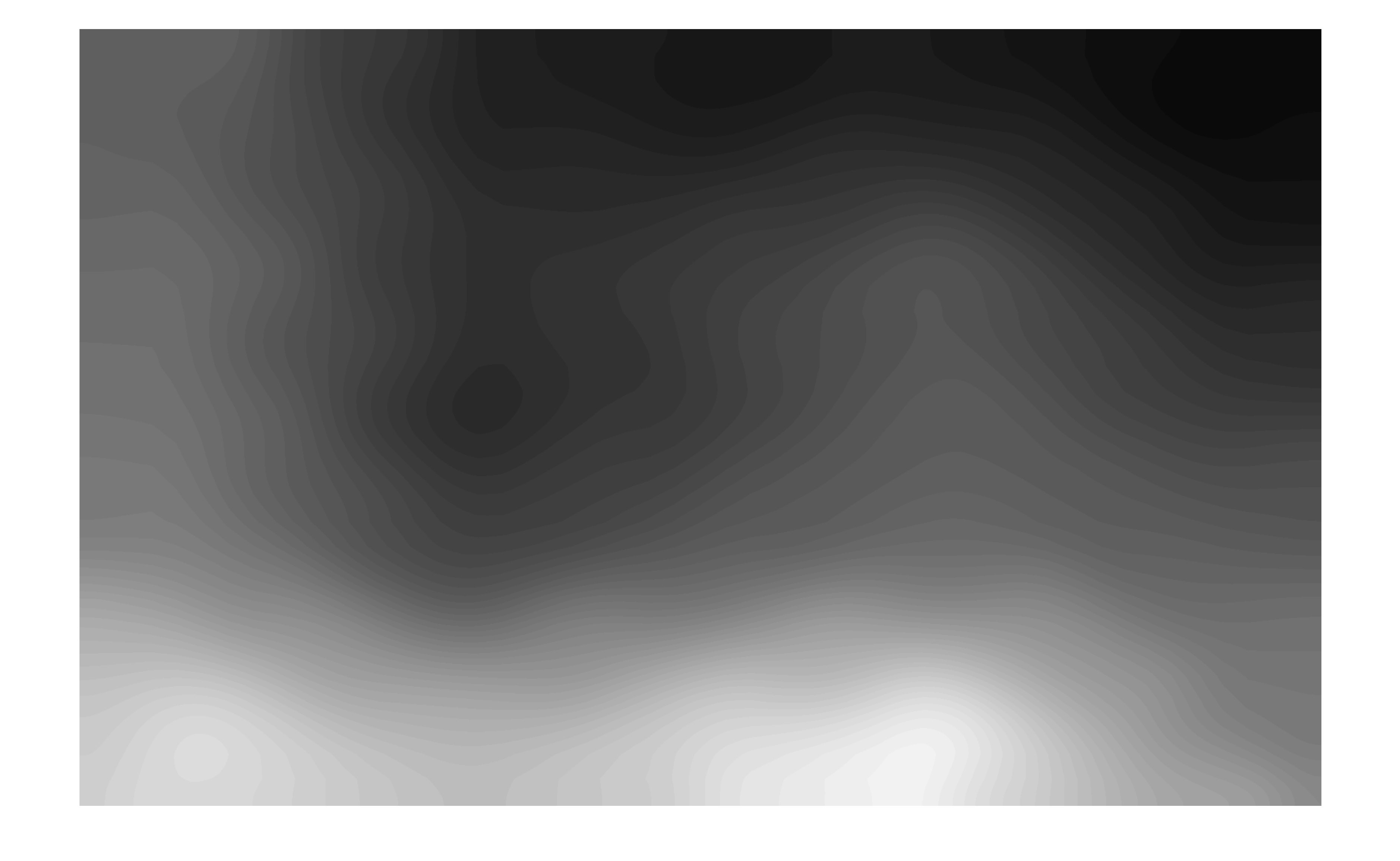}}
\subcaptionbox{$\textbf{M}_{\textbf{right~cam}}^{\textbf{L}}$ 
\label{fig_cal3}}{\includegraphics[width=0.33\columnwidth]{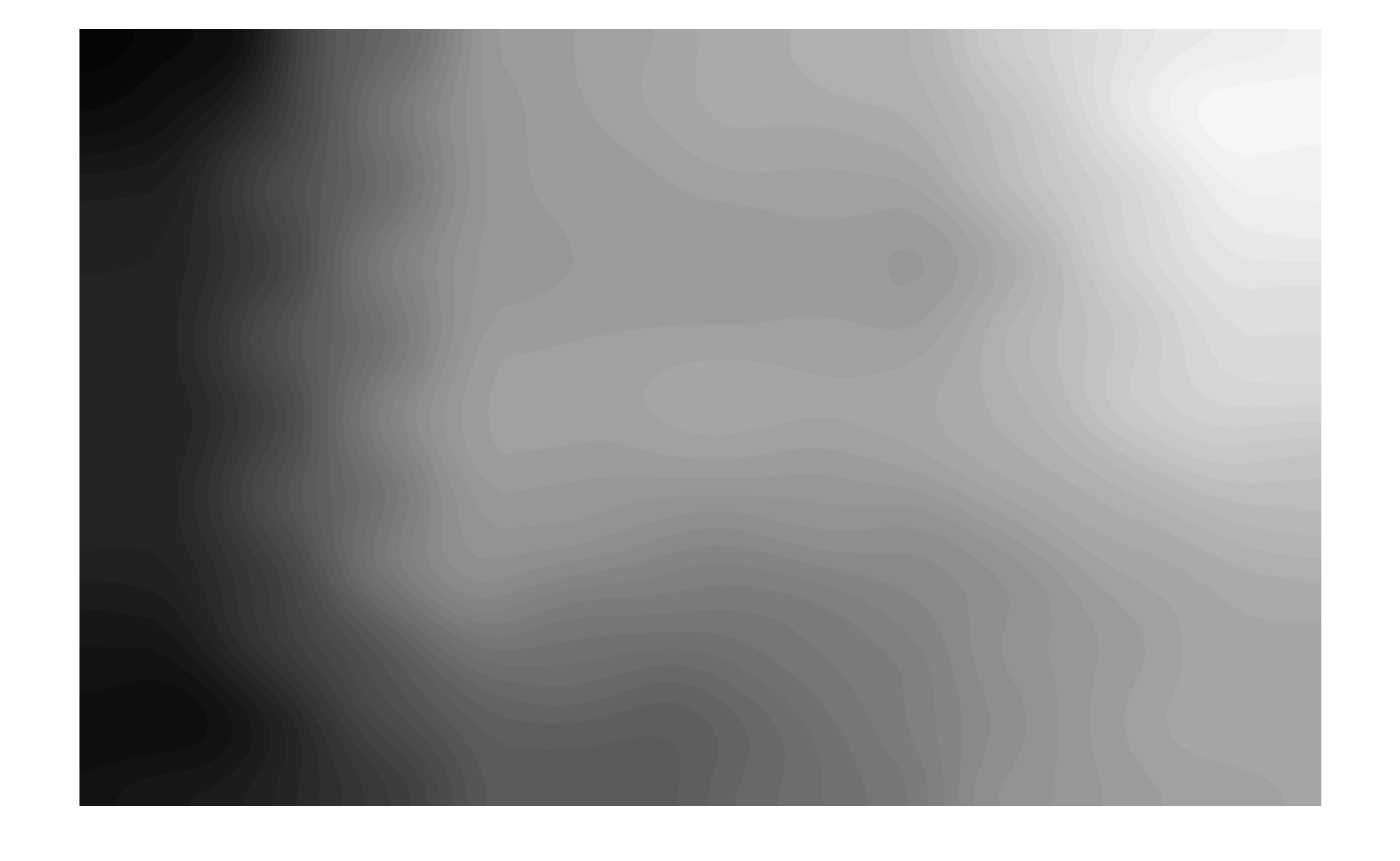}}
\subcaptionbox{$\textbf{M}_{\textbf{right~cam}}^{\textbf{R}}$ \label{fig_cal2}}{\includegraphics[width=0.33\columnwidth]{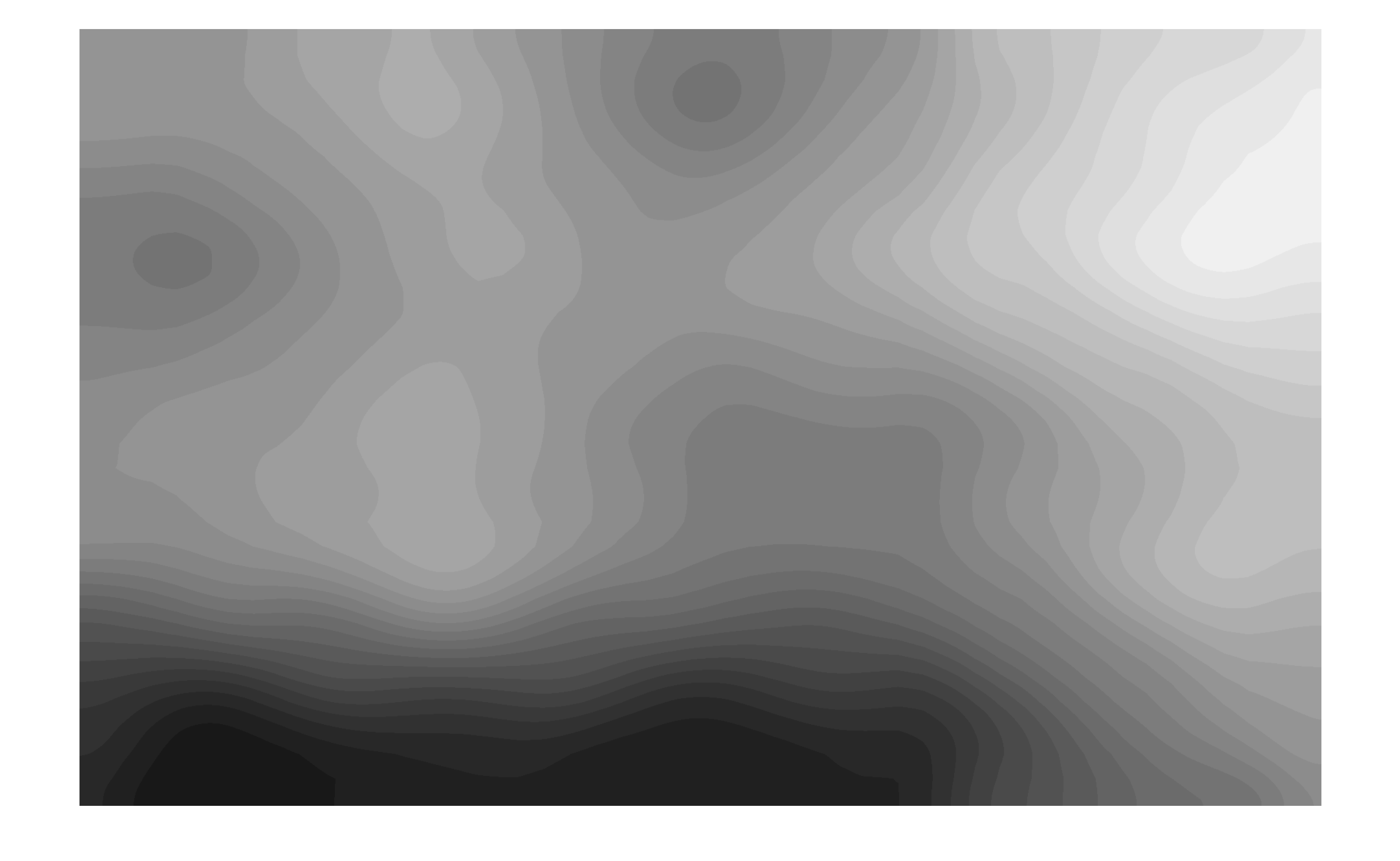}}
\vspace{-1mm}
\caption{Sample generated weight maps based on the calibration accuracy and gaze availability in a three-camera setup.}
\label{fig_maps}
\vspace{-3mm}
\end{figure*}

For generating the weight maps (${\bf M_c^e}$), we extract various statistics that are relevant to the estimation reliability (e.g., accuracy, precision, availability, gain) from the calibration data. For instance, as perhaps the most relevant and effective indicator, we calculate the estimation accuracy of each sensor (eye) on each calibration point. If a sensor's calibration accuracy around a point is consistently higher than the others, that sensor's estimation performance during testing is expected to be more reliable around the same point. Thus, higher weights are assigned to the sensors with better calibration performances for certain regions. More specifically, to calculate the weighting indicators (e.g., $acc_{c,k}^e$) for each point, after learning the calibration model on the whole calibration data, we apply the learned model on the very same data. Then, we compute the aforementioned statistics on the calibrated samples, such as the accuracy by measuring how close the calibrated samples are to their corresponding target points. We perform this process for each calibration point of each sensor independently, and obtain 2C values for each calibration point. We then normalize these accuracy values to compute the sensor weights ($w_{c,k}^e$) for each calibration point:
\begin{eqnarray}
w_{c,k}^{e} &=& \frac{acc_{c,k}^e}{\sum_c \sum_e \sum_k acc_{c,k}^e}, \label{weighting} \\
{\bf W_c^e} &=& \{w_{c,k}^{e}~|~e \in\{L,R\},~c \in\{1,2,..,C\},~1 \leq k \leq K\}, \nonumber
\end{eqnarray}
where $K$ is the number of calibration points. Lastly, we interpolate and extrapolate the weight set (${\bf W_c^e}$) over the whole screen to generate the weight maps (${\bf M_c^e}$). Sample generated weight maps from a three-camera setup are shown in Fig.~\ref{fig_maps}. In this paper, we use the estimation accuracy and availability statistics as the weighting indicators. Nevertheless, other alternatives can also be employed towards more robustly determining the sensor weights. For example, the estimation precision, which is the ability to reliably reproduce the same estimation for a target calibration point, or the histogram of the best performing sensor, which stores the information about how often each sensor achieves the best estimation for a target calibration point, could provide complementary evidence. We plan to investigate such alternative indicators in our future work. On the other hand, the main drawbacks of this method are: first, it may be sensitive to large head movements since the weights are estimated according to the calibration position, and second, an initial estimation using simple averaging is required to localize the weights on the generated maps.



\vspace{-2mm}
\section{Evaluation on Simulated Data}
\label{evaluation_simulation}
We conduct extensive simulations to primarily investigate and understand how the overall tracking performance is affected when the number of cameras is increased in various configurations. In real-world settings, the number of cameras to be employed for real-time eye tracking is limited due to cost data bandwidth constraints. Therefore, we start our evaluations on the simulated data to analyze the efficacy and limits of the proposed framework. The tracking performance is measured as the gaze accuracy and estimation availability. The accuracy is defined as the average displacement in degrees of visual angle ($^\circ$) between the stimuli points and estimated \ac{PoR}s, using all raw samples, i.e., neither temporal smoothing nor post-processing is applied. The availability is $\%$ of samples, which the system is able to compute a \ac{PoR} during the evaluation. In other words, it indicates the system's working volume.


\vspace{-3mm}
\subsection{Simulation Setup}
\label{simulation_setup}
Simulation data is generated using an open-source software framework \cite{Martin2008}. The simulator enables detailed modeling of different components of the hardware setup and an eye in 3D, and provides a realistic simulation framework. Nonetheless, simulation of non-spherical cornea, eyelid occlusions, eyewear effects, lens or sensor distortions, are currently not possible.

\begin{figure}[b!]
\centering
\vspace{-3mm}
\includegraphics[width=0.9\columnwidth]{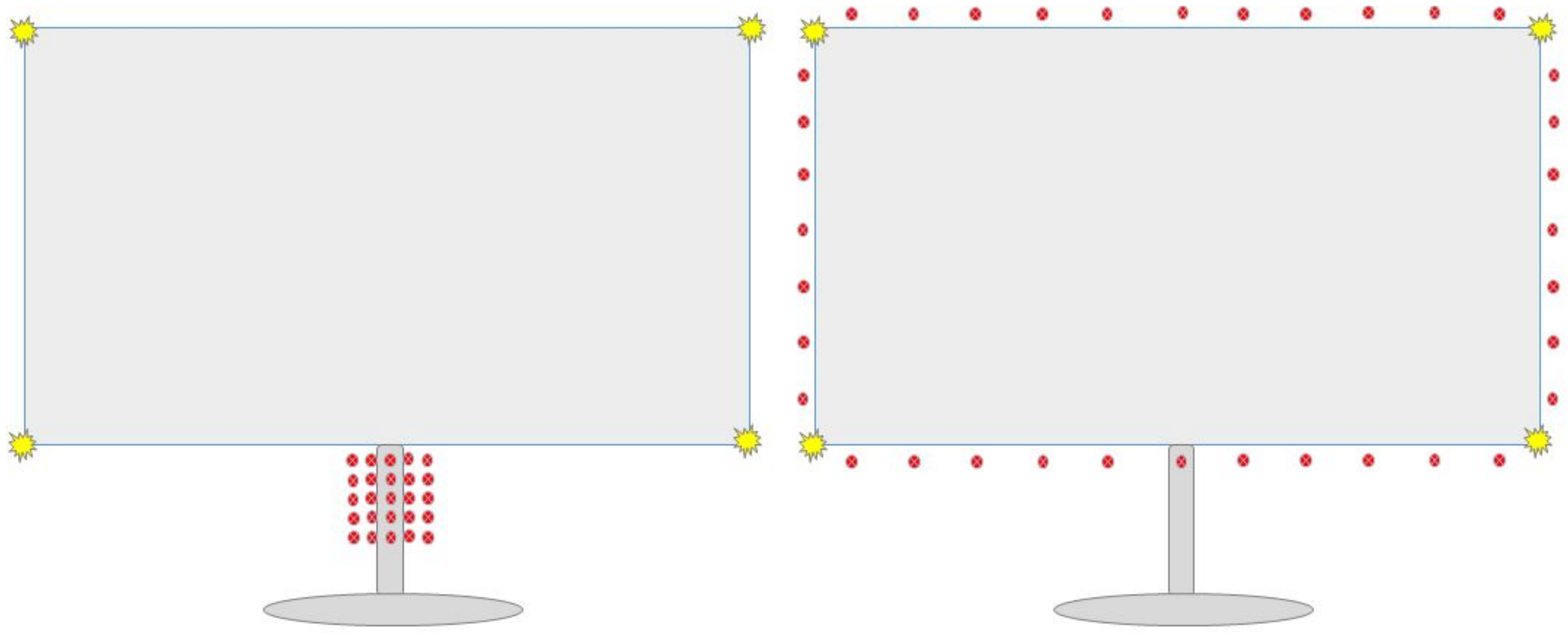}
\vspace{-2mm}
\caption{Single-view (\emph{case 0}) and multi-view (\emph{case 1}) setups.}
\label{simulations_camera_setup} 
\vspace{-3mm}
\end{figure}

\begin{figure*}[t!]
\centering
\vspace{-3mm}
\subcaptionbox{\emph{Case 0 (single-view)} \label{sim_noc_sh}}{\includegraphics[trim={0 0 0 2cm},clip,width=0.65\columnwidth]{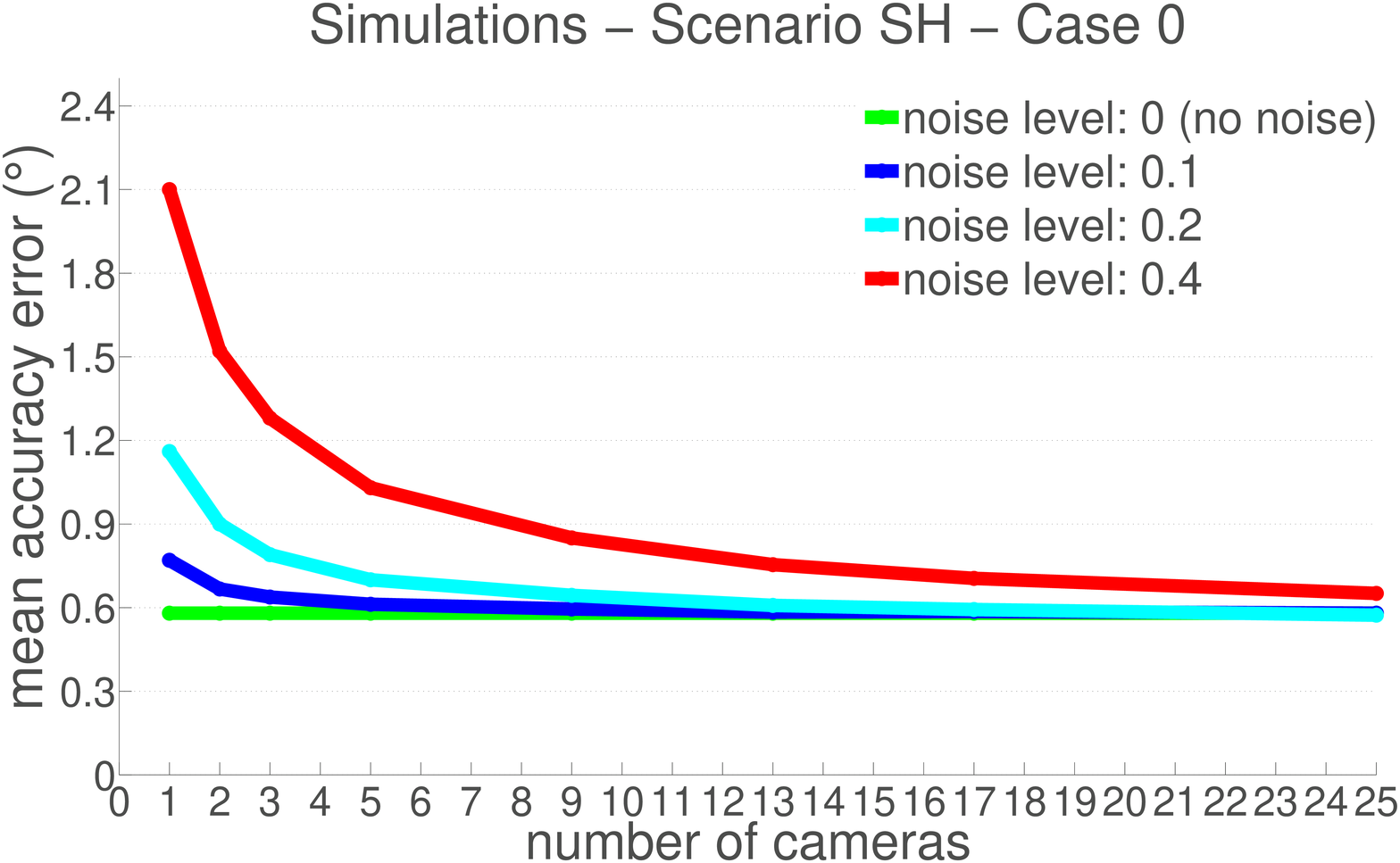}}\quad
\subcaptionbox{\emph{Case 0 vs Case 1} \label{sim_pos_sh}}{\includegraphics[trim={0 0 0 2.4cm},clip,width=0.65\columnwidth]{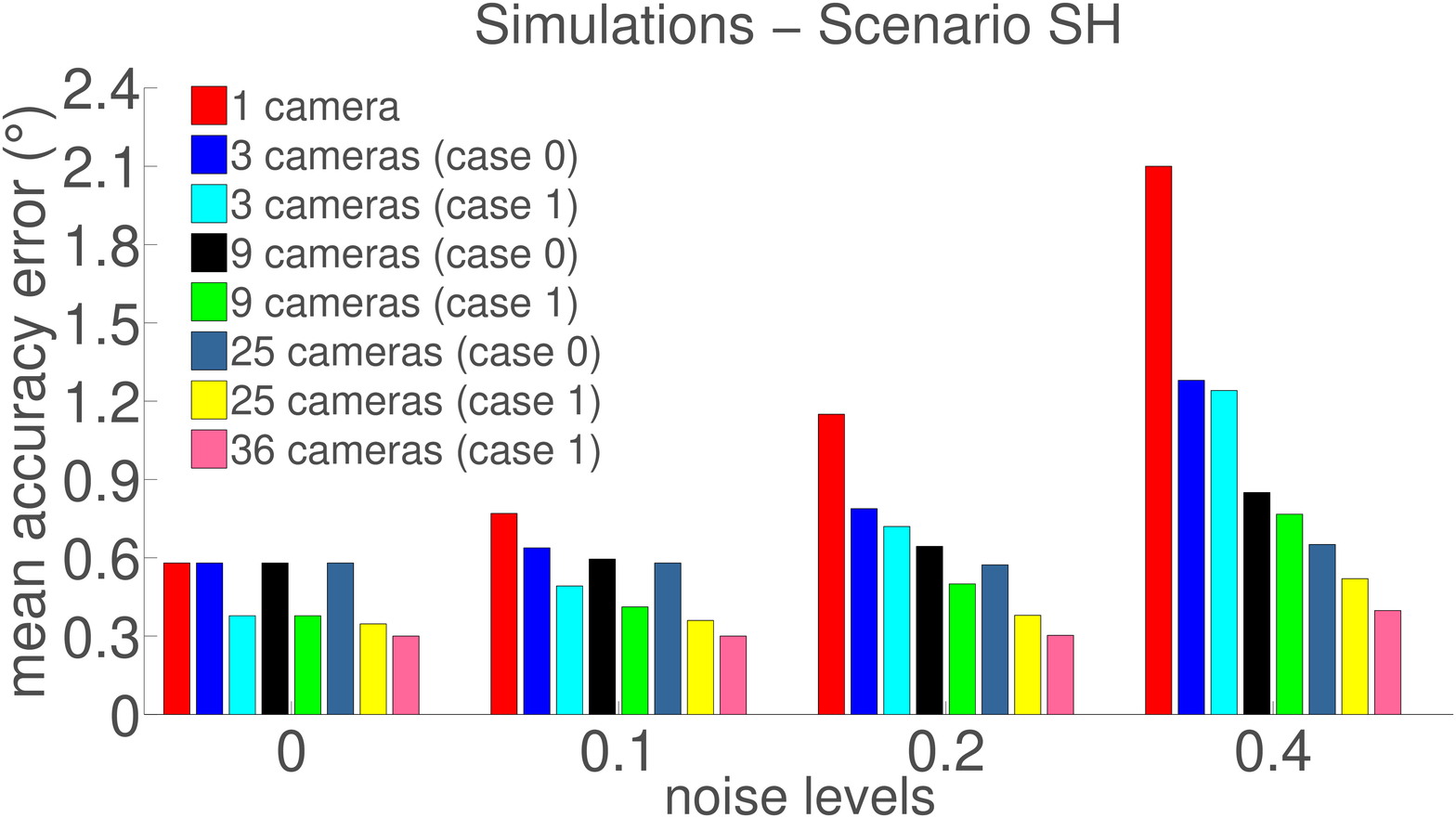}}\quad
\subcaptionbox{\emph{Case 1 (multi-view)} \label{sim_sh_comparison}}{\includegraphics[trim={0 0 0 2cm},clip,width=0.65\columnwidth]{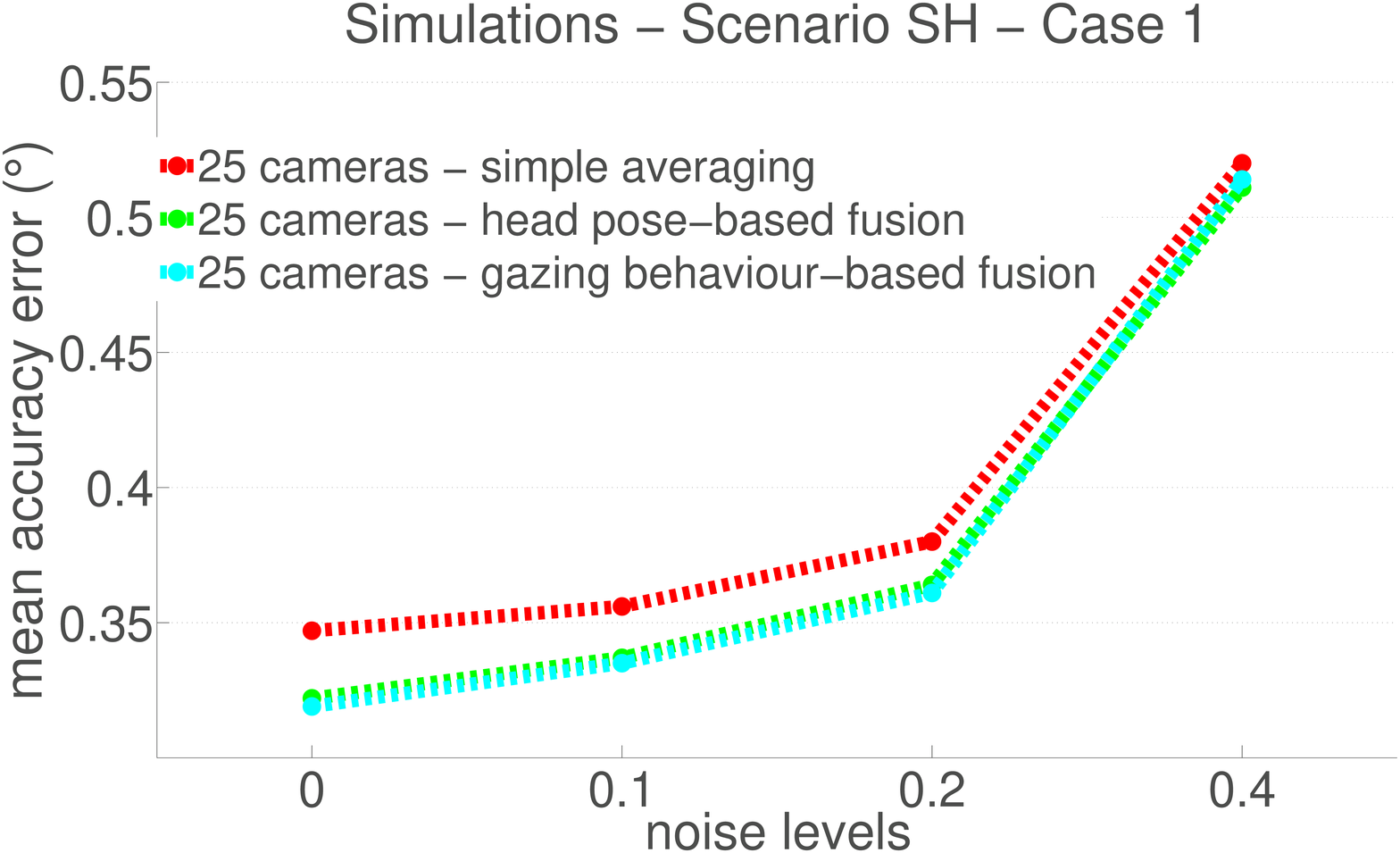}}\quad
\vspace{-1mm}
\caption{SH scenario with varying feature detection noise levels. The impact of increasing number of cameras with (a) single-view and (b) multi-view setups, and (c) a comparison of the investigated fusion methods for multi-view tracking.}
\label{fig_simulation_sh}
\vspace{-2mm}
\end{figure*}
\begin{figure*}[t!]
\centering
\vspace{-2mm}
\subcaptionbox*{}{\includegraphics[trim={0 0 0 2cm},clip,width=0.65\columnwidth]{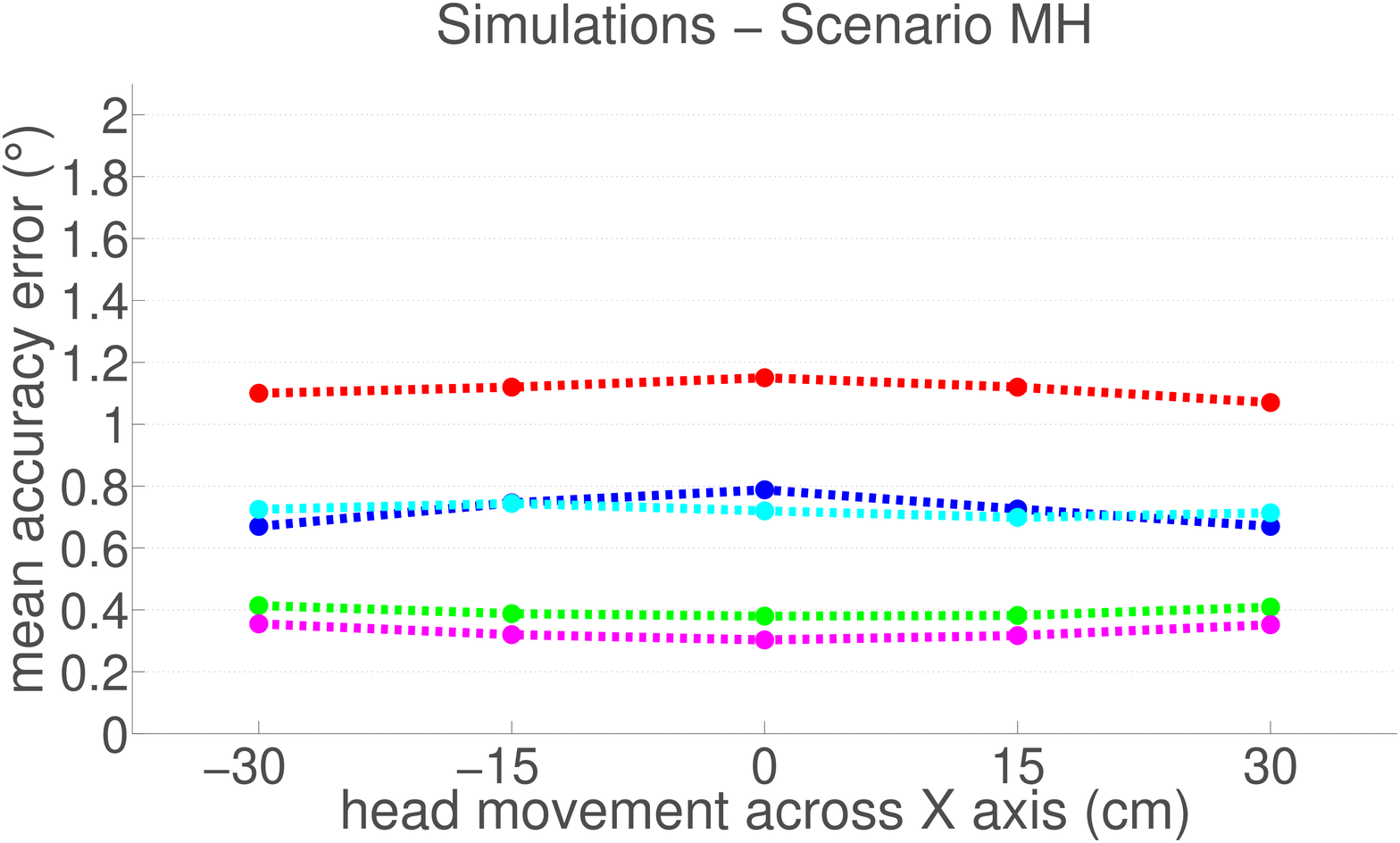}}\quad
\subcaptionbox*{}{\includegraphics[trim={0 0 0 2cm},clip,width=0.65\columnwidth]{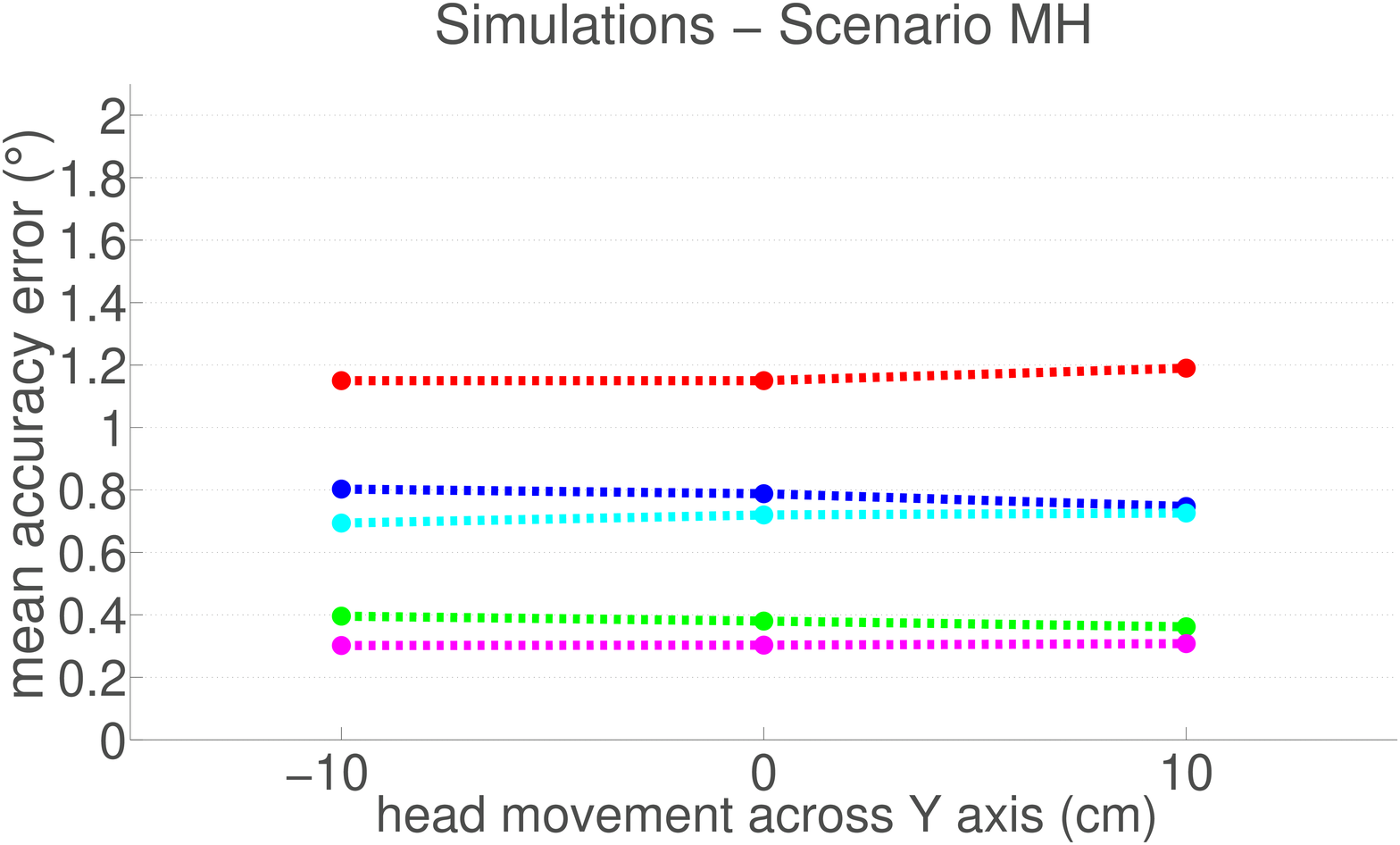}}\quad
\subcaptionbox*{}{\includegraphics[trim={0 0 0 2cm},clip,width=0.65\columnwidth]{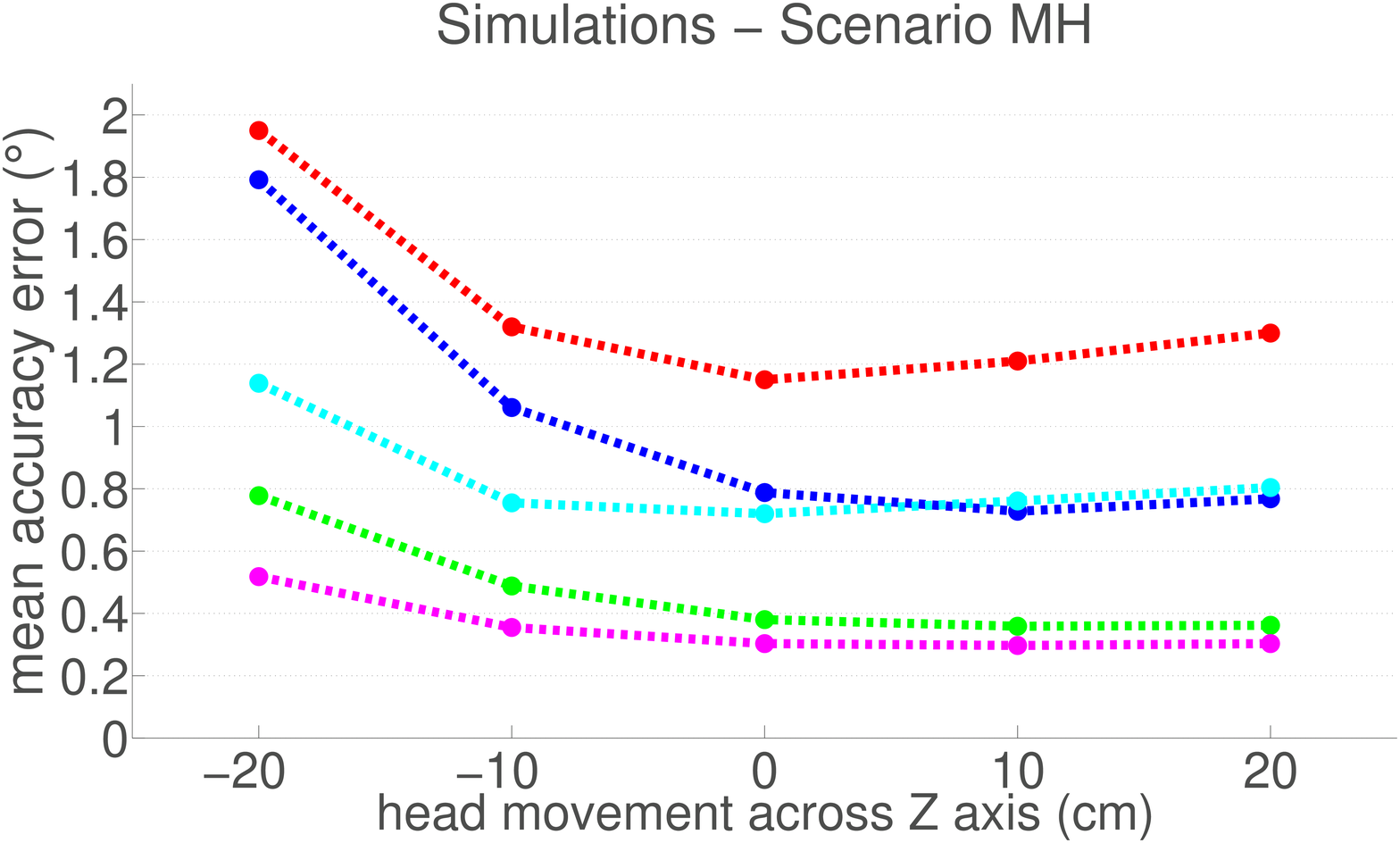}}\vspace{-6mm}
\subcaptionbox*{}{\includegraphics[trim={0 0 0 2cm},clip,width=0.65\columnwidth]{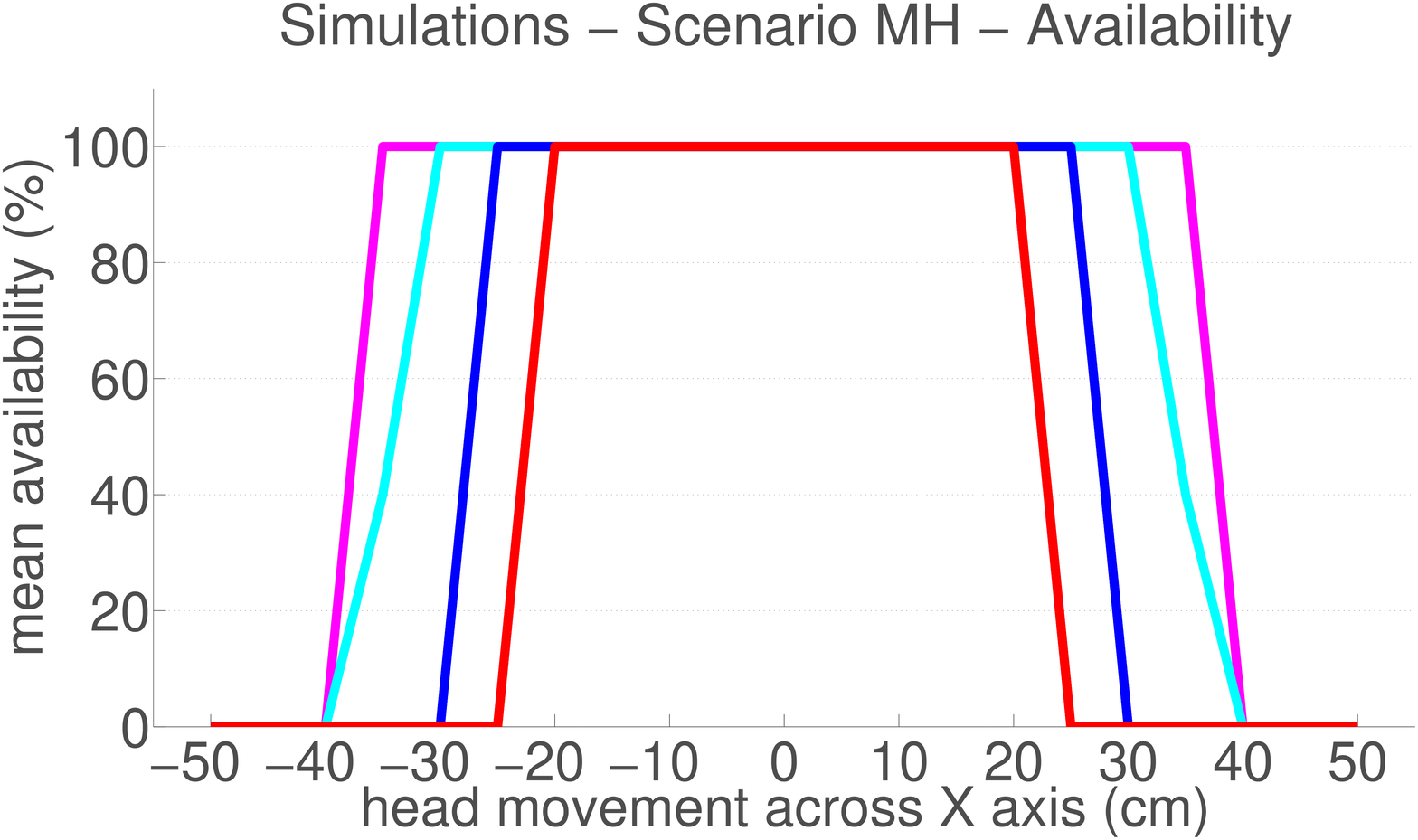}}\vspace{-3mm}\quad
\subcaptionbox*{}{\includegraphics[trim={0 0 0 2cm},clip,width=0.65\columnwidth]{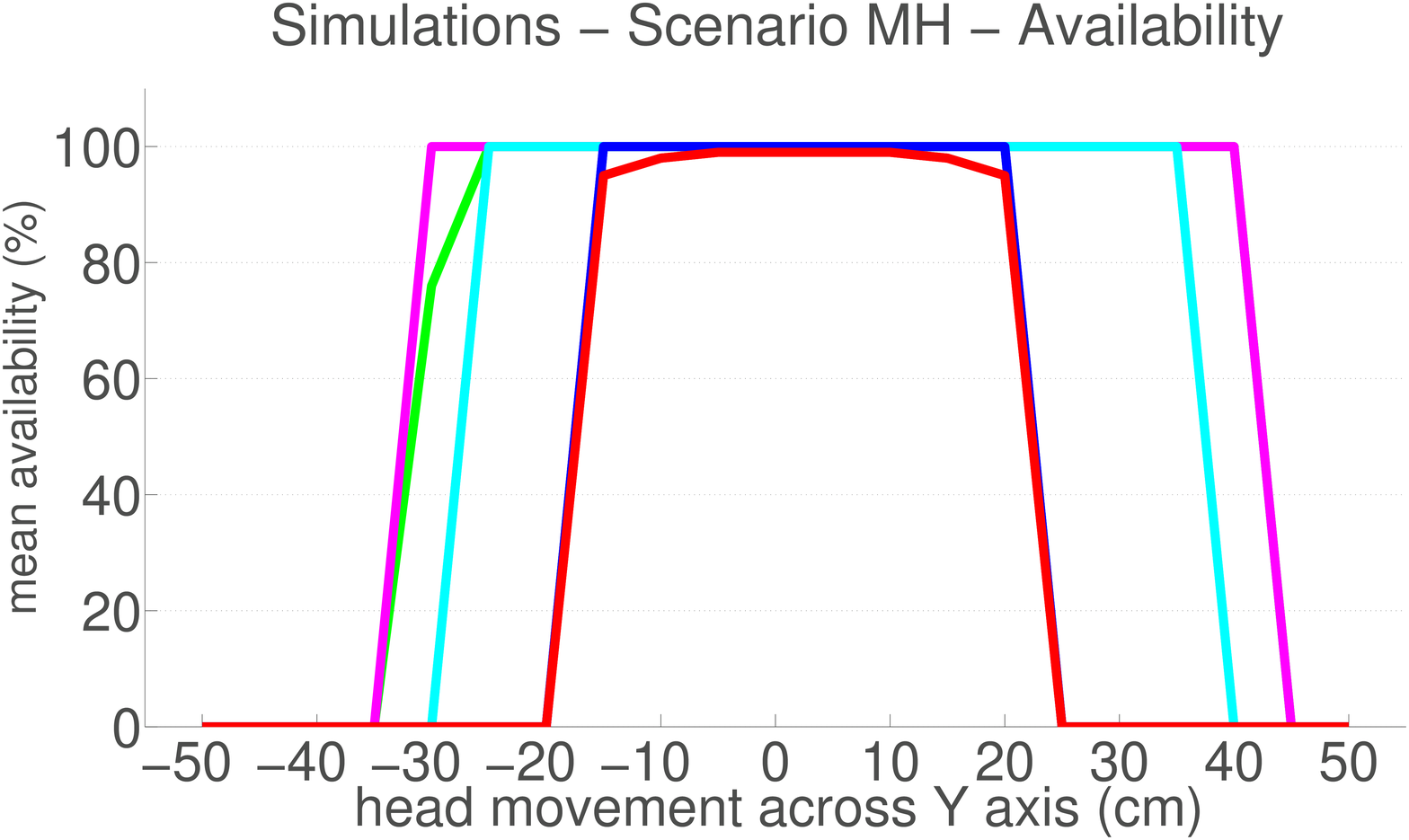}}\quad
\subcaptionbox*{}{\includegraphics[trim={0 0 0 2cm},clip,width=0.65\columnwidth]{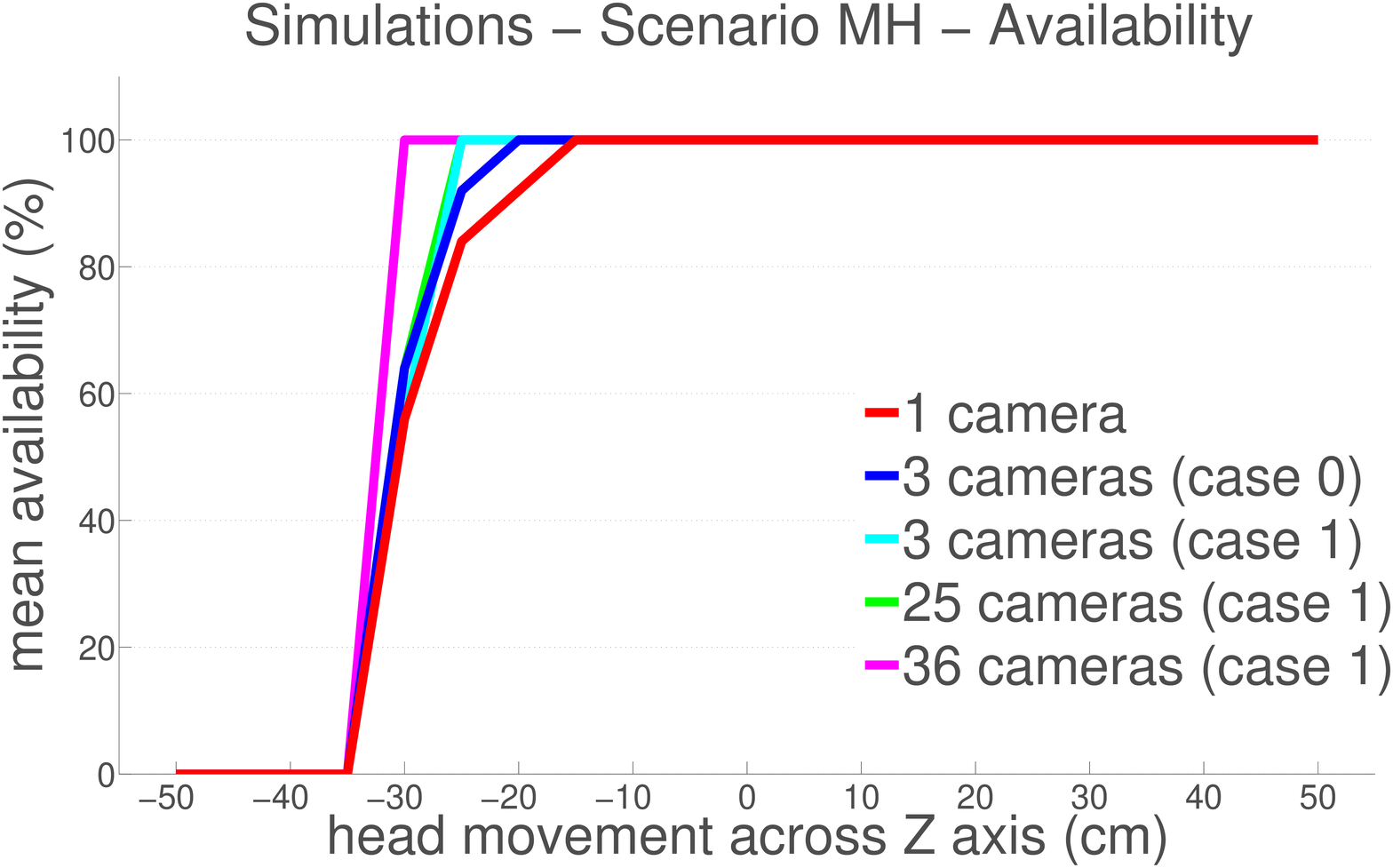}}\quad
\vspace{-3mm}
\caption{MH scenario with real-world (0.2) noise level. The impact of number of cameras and their configurations (\emph{case 0} and 1) on the head movement robustness (top row) and gaze availability (bottom row) when user moves from the default calibration position (0, 20, 60) along X, Y, and Z directions. This figure best viewed in color. All subfigures have the same legend.} 
\label{fig_simulation_mh}
\vspace{-3mm}
\end{figure*}

\begin{figure}[b!]
\vspace{-8mm}
\centering
\includegraphics[width=0.7\columnwidth]{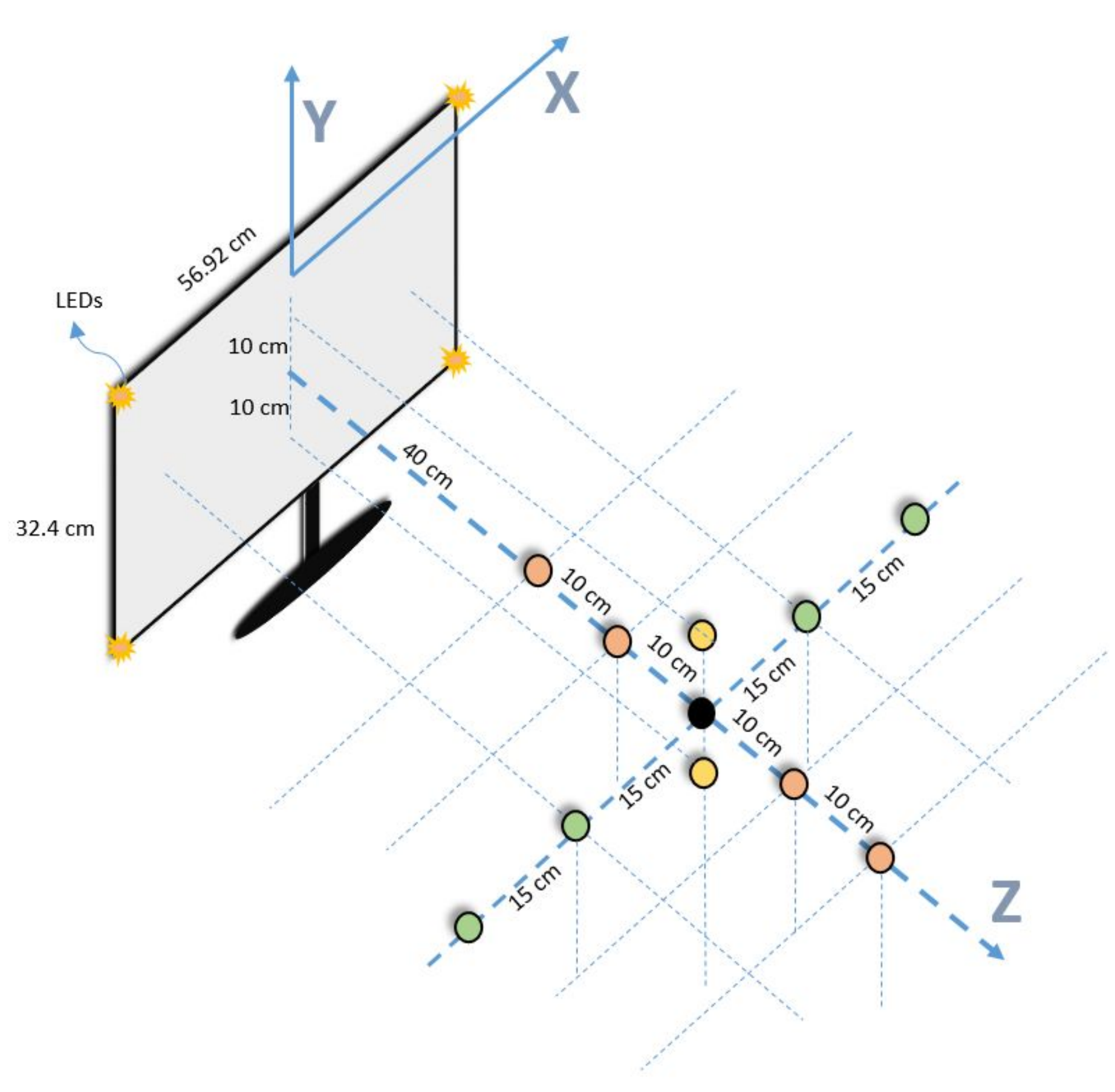}
\vspace{-4mm}
\caption{Simulation setup. Calibration is at the black circle.}
\label{simulations_setup}
\end{figure}

To simulate the impact of increasing the number of cameras, we setup two configurations: i) single-view tracking by placing multiple cameras densely at the bottom of a screen (\emph{case 0}) and ii) multi-view tracking by placing them uniformly around a screen (\emph{case 1}), as visualized in Fig.~\ref{simulations_camera_setup}. We simulate an eye using the typical eye parameters listed in \cite{Guestrin2006}. In addition, we simulate a similar environment that we have in the user experiments, i.e., 24-inch screen, 4 light sources, cameras with 1280$\times$1024 pixels resolution, and lenses with 8mm focal length (diagonal \ac{FoV}=58$^\circ$) to allow for large head movements. We acquire calibration and test data using the simulated environment. For the calibration, we generated data when gazing at 9 uniformly distributed target stimuli points on the screen, whereas for the test data acquisition, we randomly generate 18 test points in order to avoid reporting over-optimistic results due to overfitting on the calibration point locations. The test points are displayed in a 3$\times$3 grid with 2 points per region to cover the whole screen. In addition, to simulate realistic test conditions, we alter the noise level to examine the impact of noise-free and noisy data. For each point, we collect 100 samples, and introduce uniformly distributed feature position errors with a maximum magnitude of 0.4 pixels per feature (noise level $\in\{0, 0.1, 0.2, 0.4\}$).

The simulations are performed under two different scenarios, namely, \emph{Stationary Head (SH)} and \emph{Moving Head (MH)}. In \emph{SH} scenario, the eye is located 60 cm to the screen and kept at the same position during the experiments. Whereas, in \emph{MH} scenario, the eye location is changed along three directions, X, Y, and Z as shown in Fig.~\ref{simulations_setup}. In both scenarios, the calibration is performed at the default head position (0, 20, 60) cm.

\vspace{-3mm}
\subsection{Results on Stationary Head (SH) Scenario}

In \emph{SH} scenario, the main emphasis is given to the impact of increased number of cameras on estimation accuracy as there is no head movement. Fig.~\ref{fig_simulation_sh} shows the obtained results under various setup configurations while also altering feature detection noise amount to understand the theoretical and practical impact of setup configurations. In \emph{case 0} (single-view tracking), when no noise is introduced, increasing the number of cameras, even up to 25 cameras, does not provide any estimation accuracy improvement (see Fig.~\ref{sim_noc_sh}). Contrarily, when a significant amount of noise is introduced, the more cameras the system employs, the higher accuracies it achieves since the noisy outputs are smoothed out.

The simulation results shown in Fig.~\ref{sim_pos_sh} indicate that not only the number of cameras, but also their configuration is crucial for improving the tracking performance. For instance, when there is no noise, a 3-camera multi-view setup (\emph{case 1}) outperforms a 25-camera single-view setup (\emph{case 0}). In addition, the results suggest that when higher levels of noise (e.g., 0.4) are introduced, the number of cameras has more impact than the configuration of cameras since more cameras can better filter the noise out. Under the real-world noise level ($\sim$0.2), the camera configuration is undoubtedly more effective than the number of cameras. In other words, the multi-view configuration always outperforms the single-view one when employing the same number of cameras. Furthermore, Fig.~\ref{sim_sh_comparison} shows the comparison of non-adaptive and adaptive fusion mechanisms. The results indicate that the proposed adaptive mechanisms perform better than the simple averaging. 





\vspace{-3mm}
\subsection{Simulation Results on Moving Head (MH) Scenario}
In \emph{MH} scenario, we examine the impact of single- and multi-view setups, particularly on the estimation availability and head movement robustness along X, Y, and Z directions. For this scenario, the virtual eye is calibrated at the default position (0, 20, 60) and real-world noise level (0.2) is introduced. The eye then is moved to various locations along three directions as shown in Fig.~\ref{simulations_setup} and the tracking is performed in these locations using the learned calibration at the default position. As depicted in Fig.~\ref{fig_simulation_mh} (top row), the tracking is highly robust to head movements along X (horizontal) and Y (vertical) directions. In fact, even the single-camera system is highly tolerant owing to the employed gaze estimation method and user calibration technique \cite{Arar2016j}. In these cases, increasing the number of cameras simply enhances the overall estimation accuracy. On the other hand, the robustness to head movements along Z axis (depth translations) is very challenging for \emph{cross ratio-based} systems due to insufficient bias correction. As the calibration is learned as an offset at a fixed head location, the learned offset does not sufficiently compensate for the bias when the user moves away from the calibrated position, especially along Z axis. Therefore, such movements cause a significant decay in estimation accuracy for a single-camera system. Nonetheless, as clearly depicted from the line slopes in Fig.~\ref{fig_simulation_mh} (top-right), multi-view setups (\emph{case 1}) yields a significant tolerance compared to single-camera or single-view (\emph{case 0}) setups. For instance, 3-camera multi-view configuration improves the accuracy by 43\% and 35\% in comparison to single-camera and 3-camera single-view configuration, respectively.



Furthermore, an important benefit of the multi-view tracking is the increased estimation availability. Fig.~\ref{fig_simulation_mh} (bottom row) demonstrates the impact of setup configurations on the gaze availability, in \%, when the user moves along X, Y, and Z directions. The results clearly show that multi-view setups allow for significantly larger head movements (working volume) in all three directions in comparison to single-view setups. For instance, 3-camera multi-view setup (\emph{case 1}) provides an additional $\pm$15 cm and $\pm$10 cm head movement tolerance along X and Y directions, respectively. The reason is that each camera has a different \ac{FoV}, and consequently, the overall \ac{FoV} increases with the fusion of all \ac{FoV}s. Note also that increasing the number of cameras from 3 to 36 does not drastically improve the availability as their \ac{FoV}s starts to overlap.


\vspace{-6mm}
\section{Evaluation on User Experiments}
\label{evaluation_real}
This section describes the evaluation of our approach on real-world data obtained through user experiments. 

\vspace{-3mm}
\subsection{Hardware Setup}
\label{hardware_setup}
\begin{figure}[t!]
\vspace{-3mm}
\centering
\includegraphics[width=0.8\columnwidth]{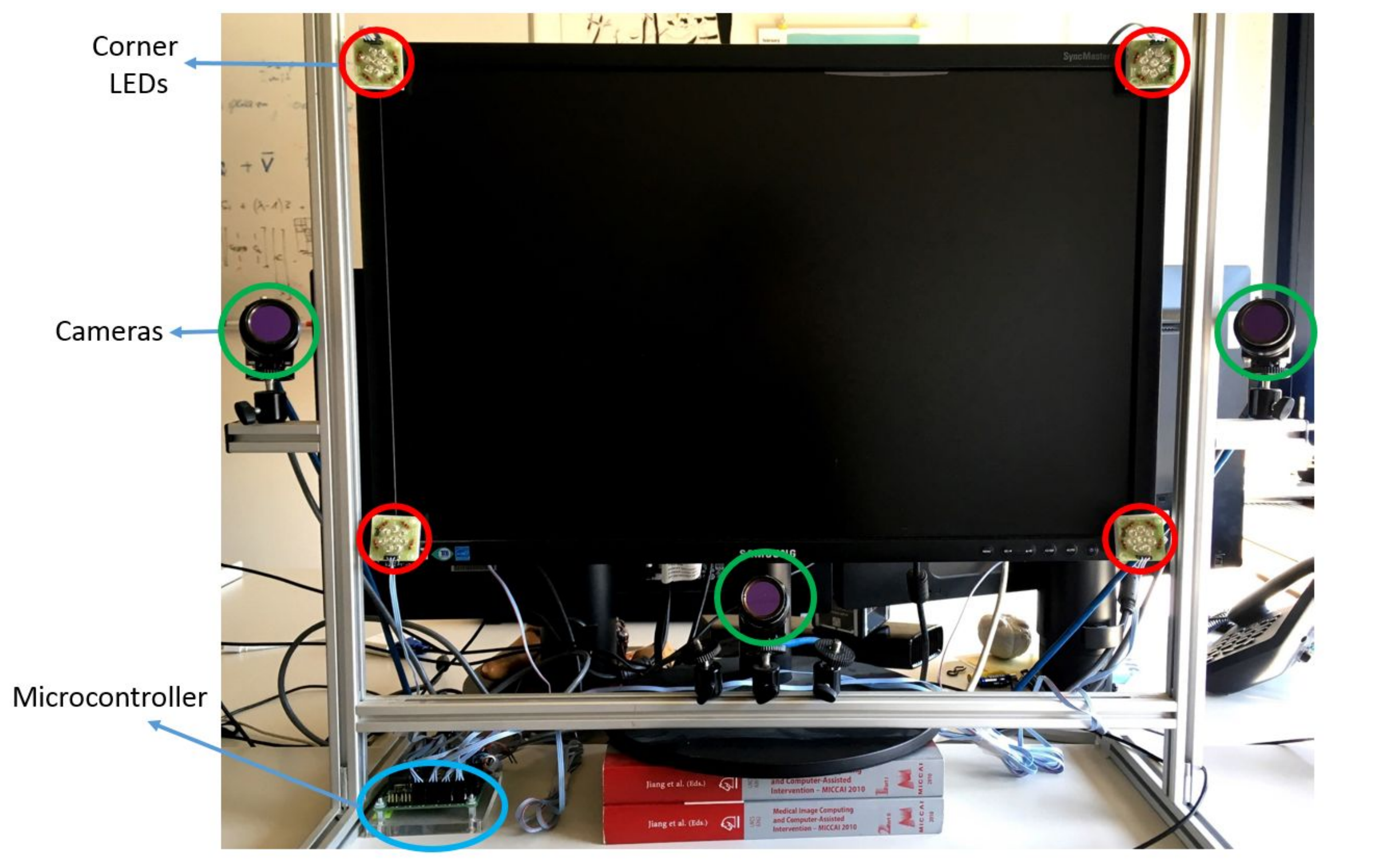}
\vspace{-1mm}
\caption{Proposed 3-camera prototype setup.}
\label{hardware} 
\vspace{-5mm}
\end{figure}

Our prototype setup consists of three PointGrey Flea3 monochrome cameras, four groups of \ac{NIR} LEDs for the illumination, and a controller unit for the synchronization. Each camera has an image resolution of 1280$\times$1024, and is equipped with an 8 mm manual focus lens (diagonal \ac{FoV}=58$^\circ$). The cameras are installed on a frame around a 24-inch monitor as shown in Fig.~\ref{hardware}. One camera is located slightly below the screen, whereas the other two are placed on the left and right sides of the screen. Note that the setup corresponds to the 3-camera multi-view (\emph{case 1}) configuration described in the simulations. In order to create the glints, 850nm-wavelength LEDs are placed on the corners of the monitor. A micro-controller is programmed to synchronize all cameras, so that the images are simultaneously captured from all cameras at 30 fps. In addition, we optimize the light emissions regarding the eye safety by synchronizing the cameras' shutters with the emission duration of LEDs.

\vspace{-3mm}
\subsection{Dataset \& Experimental Protocol}
\label{database}
A series of user experiments were conducted to comprehensively evaluate the proposed framework regarding the estimation accuracy, availability, robustness against head movements, varying illumination, use of eyewear, and between-subject variations. In total, 20 subjects, most of whom had no previous experience with any gaze tracking system, participated in our user experiments. 11 participants did not have any eyewear, while 5 and 4 participants wore glasses and contact lenses, respectively. The participants are from diverse origin. Therefore, eye shapes and appearances exhibit a large variability.

Each participant was asked to follow 8 different experiments as described in Table~\ref{real_setup}. Experiment $\#$2 being the default protocol, in the first three experiments, we analyzed the system's tolerance to varying ambient illumination conditions, i.e., sunlight, darkness, indoor light. The remaining five experiments were designed to evaluate the system's robustness to head movements. Four of them were conventional experiments, in which the subjects were asked to move along X (horizontal) and Z (depth) axes. The remaining one stood for a novel scenario, in which the subjects were asked to continuously move their head while still fixating on the displayed gaze points. The goal of this experiment was to analyze the system's sensitivity to continuous head movements, head pose changes, and slight head translations during the fixations. In fact, this represents better the real-world scenarios, e.g., free-head gazing while listening music or talking on the phone. As our evaluation targeted natural human-computer interaction, we aimed to collect the ground truth data as natural for the subjects as possible. For instance, chinrest was avoided to keep the subject's head still and to keep the eye within the cameras' \ac{FoV} to capture high-resolution eye data, as frequently performed by previous work. In addition, the subjects were asked to gaze at the target stimuli points in a natural and comfortable way. As a result, the subjects had different head-pose and eye-pose characteristics, facial expressions, and heights (along Y axis) while gazing. Data acquisition and performance evaluations are done similar to the simulations described in Section~\ref{simulation_setup}. The default user-to-screen distance is set to 60 cm and user calibration is performed only at this distance. The learned calibration is then applied during testing for all configurations. Head pose statistics, sample video frames from each experiment, and an example video (Exp \#2 vs Exp \#3) are provided in the supplementary material.

\begin{table}[t!]
\vspace{-3mm}
\caption{Experimental configurations.}
\vspace{-2mm}
\label{real_setup}       
\centering
\begin{tabular}{ccccl}
\hline\noalign{\smallskip}
Exp. &Lighting &  \multicolumn{2}{c}{Head (X, Z)}& Experimental Variable\\
\noalign{\smallskip}\hline\noalign{\smallskip}
0 & sunlight & 0		& 60		& illumination \\ 
1 & darkness   & 0		& 60 		& illumination  \\ 
2 & indoor & 0 		& 60 		& illumination \\ 
3 & indoor & 0		& 60 		& continuous head movements (HM)  \\
4 & indoor & 0		& 50 		& -10 cm HM along Z axis  \\
5 & indoor & 0		& 70 		& +10 cm HM along Z axis  \\
6 & indoor & +15    & 60 		& +15 cm HM along X axis  \\
7 & indoor & -15	& 60 		& -15 cm HM along X axis  \\
\noalign{\smallskip}\hline
\end{tabular}
\vspace{-4mm}
\end{table}

\begin{figure*}[t!]
\vspace{-3mm}
\centering
\subcaptionbox{Horizontal movements (along X axis)\label{ue_mh_x}}{\includegraphics[trim={0 0 0 3cm},clip,width=0.65\columnwidth]{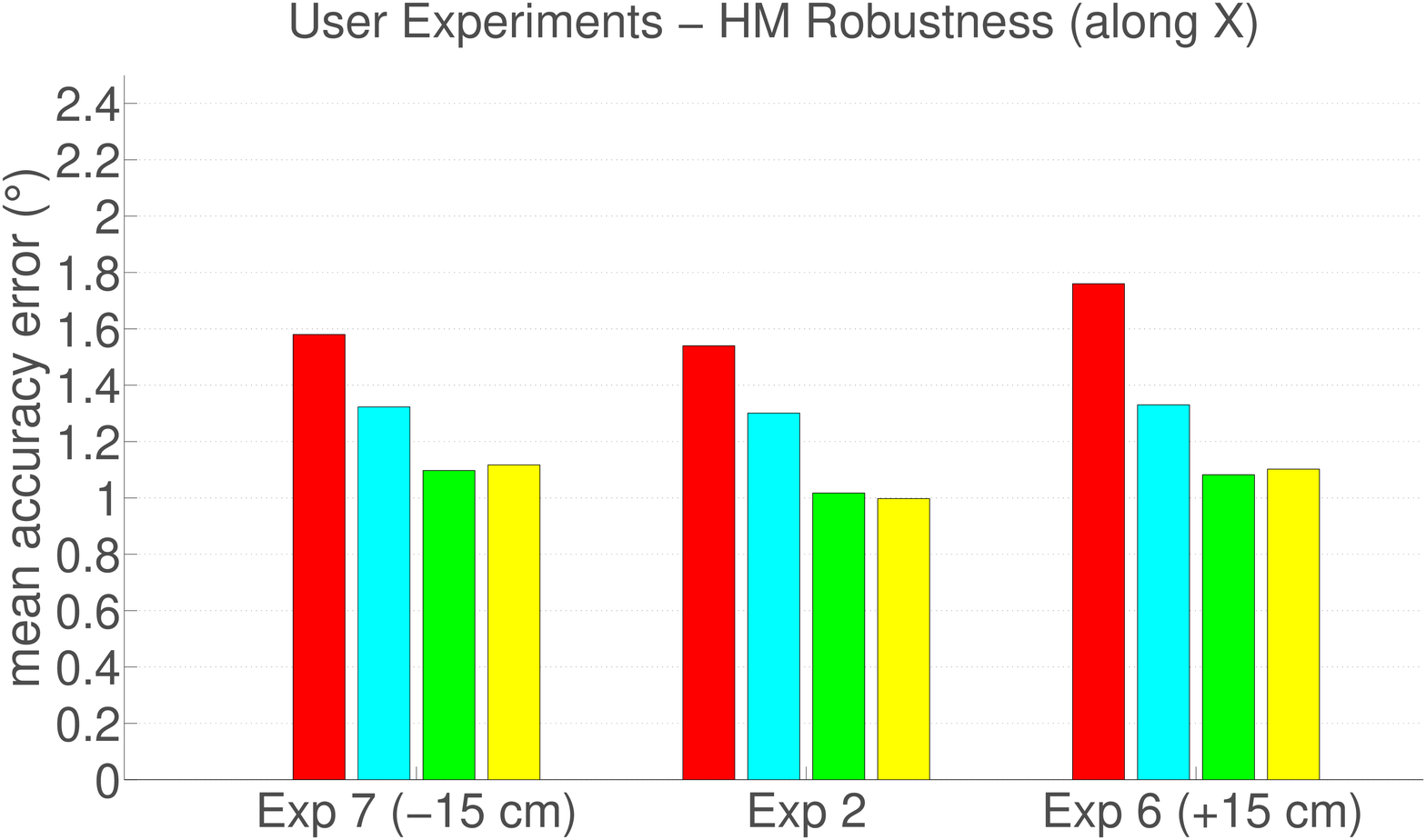}}\quad
\subcaptionbox{Depth movements  (along Z axis)\label{ue_mh_z}}{\includegraphics[trim={0 0 0 3cm},clip,width=0.65\columnwidth]{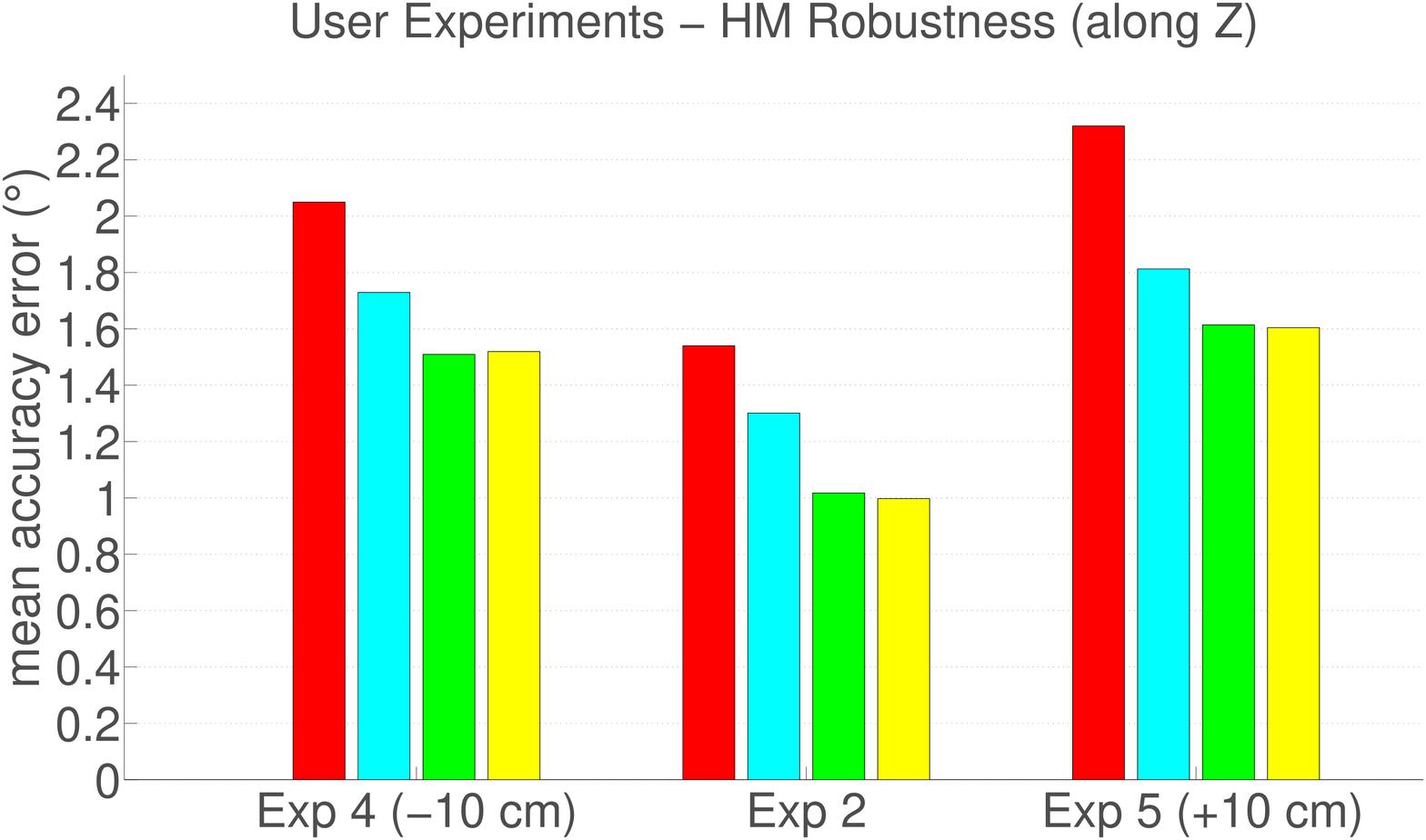}}\quad
\subcaptionbox{Continuous movements\label{ue_mh_chm}}{\includegraphics[trim={0 0 0 3cm},clip,width=0.65\columnwidth]{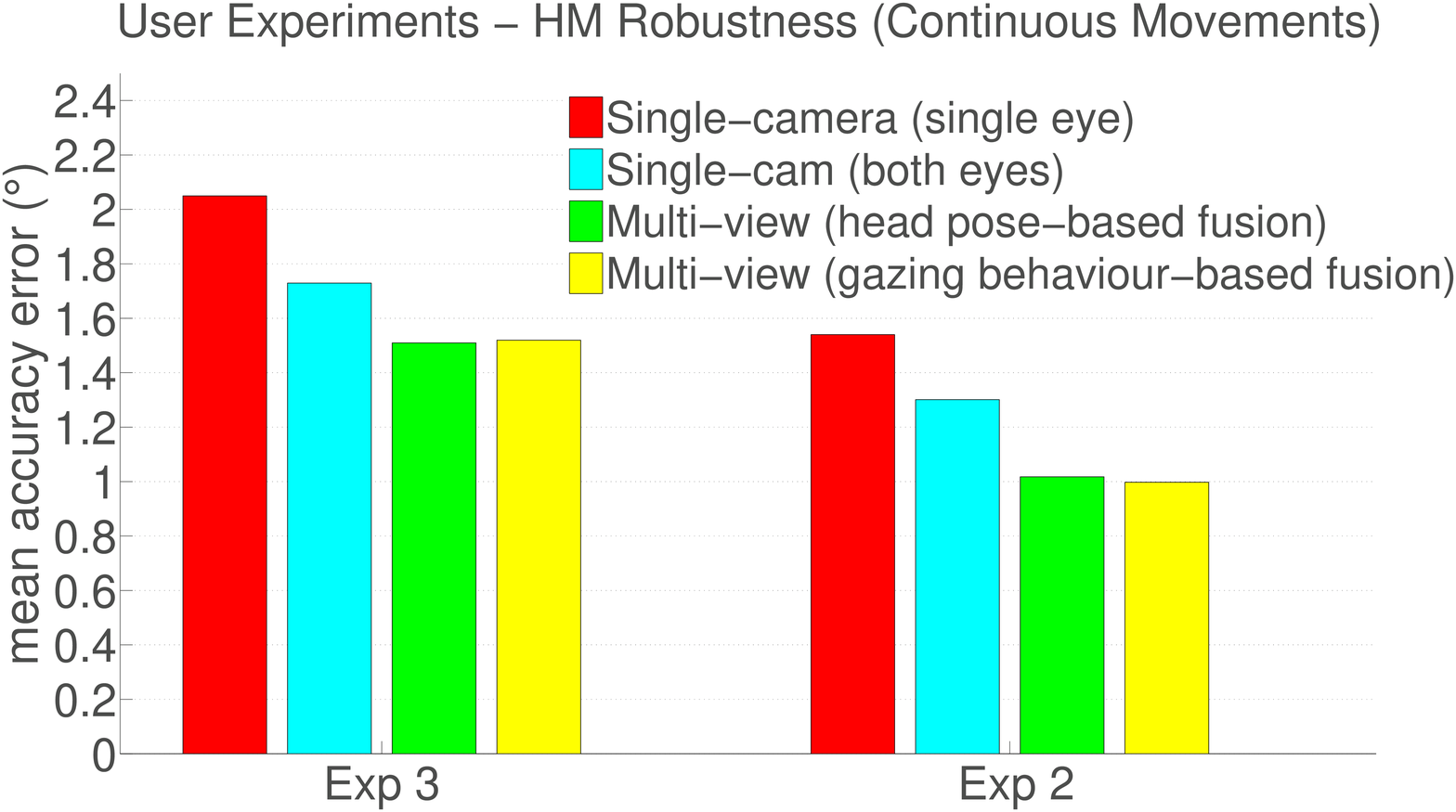}}\quad
\vspace{-2mm}
\caption{Performance comparison of single- and multi-view setups under varying head movement scenarios.}
\label{fig_user_experiments_mh}
\vspace{-3mm}
\end{figure*}

\vspace{-3mm}
\subsection{Results}
\label{evaluation_real_results}
Our framework starts with face tracking on the captured frames, in which we extract eye regions of size $\sim$90$\times$50 pixels. Feature detection is then performed to detect the pupil center and four glints. The size of the polygon formed by the glints is $\sim$9$\times$5 pixels. Next, we apply \emph{cross ratio-based} gaze estimation with the detected gaze features to calculate raw \ac{PoR}s. We then apply the learned calibration models on raw gaze outputs to compensate for the person-specific bias. Lastly, calibrated \ac{PoR}s obtained from each sensor are combined using the adaptive fusion mechanism to output an overall \ac{PoR}. In the following subsections, we present and discuss the results of various experiments. 


\begin{table}[b!]
\vspace{-3mm}
\caption{Comparison of single- and multi-view setups.}
\vspace{-1mm}
\label{table_single_vs_multi}
\centering
\begin{tabular}{lccc}
\hline\noalign{\smallskip}
\multirow{2}{*}{\parbox{3 cm}{Setup configuration}}	& Eye & \multicolumn{2}{c}{Estimation}\\
 	& Data & ($^\circ$) & 	(\%)\\ 
\noalign{\smallskip}\hline\noalign{\smallskip}
Single-cam left eye only & 1	        & 1.4	& 77.6 \\ 
Single-cam right eye only & 1            & 1.35 	& 69.6  \\ 
Single-cam both eyes & max 2 & 1.25  & 93.4  \\\hline 
Multi-cam (\emph{case 0}) & max 6 & 1.04 & 97.2  \\ \hline
Multi-cam (\emph{case 1}) pose-based best cam selection & max 2 & 1.17	& 100  \\ 
Multi-cam (\emph{case 1}) simple averaging & max 6 & 0.89	& 100  \\ \hline
Multi-cam (\emph{case 1}) head pose-based fusion  & max 6 & 0.76	& 100 \\ 
Multi-cam (\emph{case 1}) gazing behavior-based fusion & max 6 & 0.74	& 100  \\
\noalign{\smallskip}\hline
\end{tabular}
\vspace{-3mm}
\end{table}

\subsubsection{Single-view vs Multi-view Tracking}
Our first analysis emphasizes on the benefits of multi-view tracking (\emph{case 1}) over single-view tracking (\emph{case 0}) as implemented by the majority of the existing trackers. In this regard, we conducted experiments on a subset of our dataset (3 subjects, one subject per eyewear category) using a single-camera, 3-camera single-view (\emph{case 0}), and 3-camera multi-view (\emph{case 1}) setups. Mean estimation accuracy errors and availabilities obtained using these setups on experiment $\#$2 are shown in Table~\ref{table_single_vs_multi}. The results are clearly in line with the findings of the simulations (see Fig.~\ref{fig_simulation_sh}), and demonstrate the efficacy of the proposed multi-view framework in terms of both estimation accuracy and availability. Firstly, significant accuracy improvements, by about 41$\%$ and 29$\%$, are achieved using the proposed 3-camera multi-view setup in comparison to the conventional single-camera setup (using both eyes) and 3-camera single-view setup, respectively. In addition, the estimation availability is also increased. Yet, the availability analysis is more interesting when considering head movements and eyewear robustness in Sections~\ref{robustness_hm} and~\ref{robustness_eye_wear}. Furthermore, the results show the impacts of the adaptive fusion mechanisms. Although a simple averaging standalone achieves a significant performance improvement, employing the proposed adaptive fusion algorithms further enhances the accuracy.

\subsubsection{Head Movement Robustness}
\label{robustness_hm}
To evaluate our framework's robustness to head movements, we analyze the results on experiments $\#$2-7 on all subjects. Experiments $\#$2, $\#$6, and $\#$7 account for the horizontal movements (along X axis), whereas experiments $\#$2, $\#$4 and $\#$5 account for the depth movements (along Z axis). Note that vertical movements (along Y axis) are not explicitly experimented as the subjects were asked to freely adjust their heights for convenience. Moreover, we introduced a new experimental scenario (experiment $\#$3), in which the users were asked to perform continuous head pose/location changes while still fixating on the target points. The purpose of this experiment is to measure the system's sensitivity to sudden arbitrary changes during the user interaction, which may frequently occur in real-world conditions. Fig.~\ref{fig_user_experiments_mh} illustrates the results achieved on these experiments and their cross comparisons.

For horizontal head movement robustness, the results (Fig.~\ref{ue_mh_x}) are highly in line with the simulation results (Fig.~\ref{fig_simulation_mh}), such that the system is highly tolerant (1$^\circ$ vs 1.1$^\circ$) to head movements along X axis up to $\pm$15 cm movements. On the other hand, along Z axis (depth translations), the results partially differ from the simulation results. In simulations (Fig.~\ref{fig_simulation_mh}), the estimation accuracy is shown to be negatively affected by the depth movements.
~The same result holds for the user experiments. However, we also observe that the system's performance gets worse when users move away from the screen, which contradicts the simulation results. In fact, the main reason relates to the current hardware setup, which employs manual focus lenses. Despite the aperture adjustments to obtain a larger depth-of-field, out-of-focus still occurs when user's depth varies from the default position. This causes blurry appearances, and consequently less precisely detected features. In addition, the eye image resolution gets significantly lower when the user moves away from the camera, which causes the features to be detected less accurately. Yet, in overall, the multi-view framework provides more robustness, by about 25\% in accuracy and 10\% in availability, to depth translations compared to the single-camera system. Furthermore, continuous head movements results are shown in Fig.~\ref{ue_mh_chm}. The results indicate that the framework, as expected, experiences an accuracy drop, yet it continues to output PoRs with an acceptable accuracy ($\sim$1.4$^\circ$) under such a challenging scenario. As this new experimental scenario constitutes an essential use case in real-world eye tracking, we recommend future efforts to consider it in their evaluations. 


\begin{figure}[b!]
\vspace{-5mm}
\centering
\includegraphics[trim={0 0 0 2.5cm},clip,width=0.9\columnwidth]{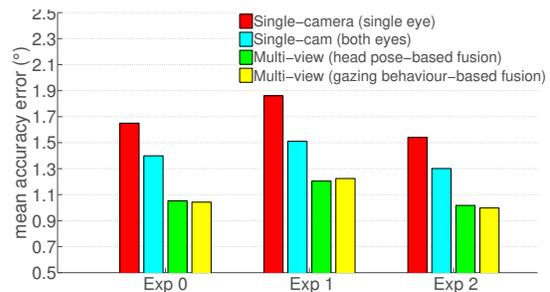}
\vspace{-3mm}
\caption{Performance comparison under varying illumination.}
\vspace{-3mm}
\label{fig_user_experiments_illumination}
\end{figure}

\begin{figure}[b!]
\vspace{-3mm}
\centering
\includegraphics[width=\columnwidth]{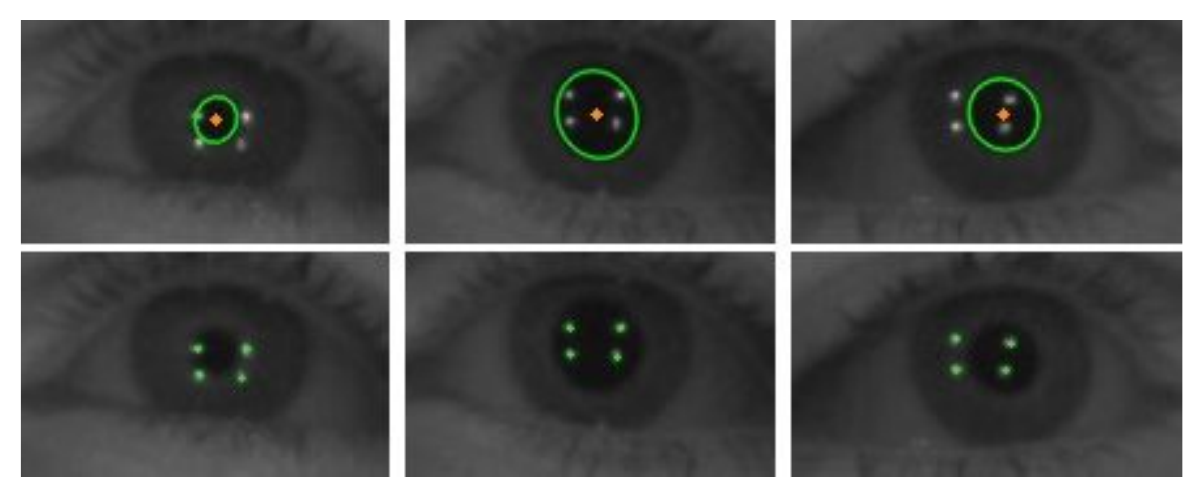}
\vspace{-6mm}
\caption{Sample appearances of eye and gaze features (glints and pupil) under varying illumination conditions: (left) sunlight, (center) darkness, and (right) indoor lighting.}
\label{illumination}
\vspace{-3mm}
\end{figure}

\subsubsection{Illumination Robustness}
Fig.~\ref{fig_user_experiments_illumination} illustrates the proposed framework's robustness under sunlight, darkness, and indoor lighting. The results indicate that ambient illumination variations do not significantly influence the estimation performance. As the suggested framework operates under active (\ac{NIR}) illumination, the system is implicitly more robust to illumination variations than natural light-based systems. For our system, the robustness is in practice more related to how robust the employed feature detection algorithms are to the changes in features when the illumination varies (Fig.~\ref{illumination}). In this respect, we proposed illumination-robust feature detection algorithms, as described in detail in \cite{ArarThesis2017}. We also note that our results under indoor lighting slightly outperforms the others because the feature detection is mainly optimized for this scenario. 


\subsubsection{Eyewear Robustness}
\label{robustness_eye_wear}
\begin{figure}[t!]
\vspace{-3mm}
\centering
\includegraphics[width=\columnwidth]{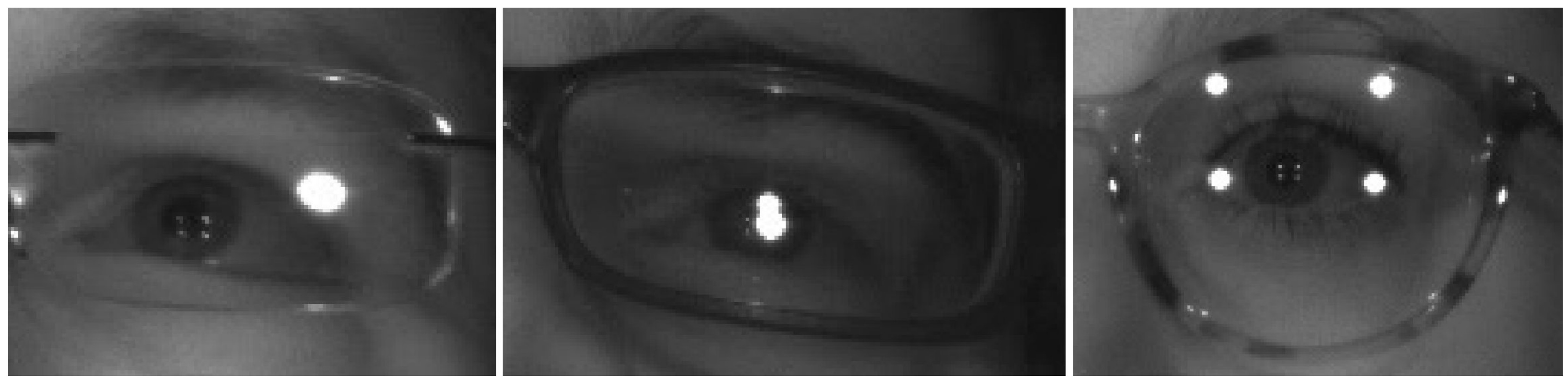}
\vspace{-6mm}
\caption{Sample impacts of glasses on eye appearance.}
\label{eye_wear_difficulties}
\vspace{-6mm}
\end{figure}

Considering that about 30$\%$ of young adults and more than half of elders in industrial nations need eyewear \cite{Schaeffel2006}, any intolerance to glasses or contact lenses undoubtedly harms the user experience. Still, it is undoubtedly one of the most challenging issues in eye tracking. Unfortunately, it has been neglected by the great majority of the previous efforts. The main challenges stem from the reflection and refraction effects on the glasses, which can significantly affect the tracking performance. Example glasses impacts on eye appearance, such as distorted features due to the refraction and coating, lost features due to the reflection, challenging feature detection due to multiple reflections, which were encountered during the user experiments can be seen in Fig.~\ref{eye_wear_difficulties}. As some of the impacts are unrecoverable, conventional single-view approaches are likely to fail under such circumstances. On the other hand, the proposed multi-view approach leverages various eye appearances per frame, in such a way to more reliably detect the features from various views. 

\begin{figure}[b!]
\centering
\vspace{-3mm}
\includegraphics[trim={0 0 0 3cm},clip,width=0.8\columnwidth]{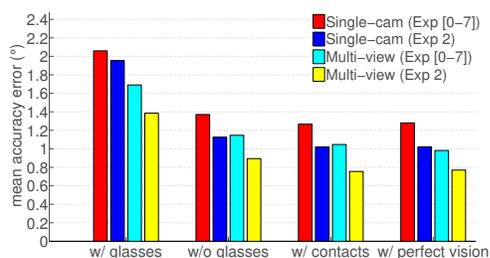}
\vspace{-3mm}
\caption{Performance comparison of varying eyewear groups.}
\label{fig_robustness_eye_wear}
\vspace{-3mm}
\end{figure}

We evaluate the efficacy of our proposed method with two separate analysis. First, we categorize the subjects into 4 groups according to their eyewear and vision quality such as the ones who wear glasses, who wear contact lenses, who do not wear glasses, and who have perfect vision. Fig.~\ref{fig_robustness_eye_wear} shows the performance comparison across these groups. The results clearly depict the improvements achieved using the multi-view setup for both the generic scenario (experiment $\#$2) and over all scenarios (experiments $\#$0-$\#$7). Among all groups, the best performance ($\sim$0.8$^\circ$) is obtained on the subjects with perfect vision (6 subjects) and contact lenses (4 subjects). As the group who do not wear glasses (15 subjects) includes some subjects with lower vision quality (who actually needs slight vision correction), a small accuracy drop is observed. Lastly, the subjects with glasses (5 subjects) achieves a lower accuracy (1.38$^\circ$ with 91.9$\%$ availability) in comparison with the other groups. However, the performance improvement, by about 0.6$^\circ$ and 10$\%$, compared to the single-camera setup highlights the benefits of the multi-view approach.




Furthermore, to eliminate between-subject variations, we compare the tracking performance on the same subject, who completed the user experiments firstly by wearing glasses and then once again by wearing contact lenses. The results shown in Table~\ref{contact_vs_glasses_murat} shows the efficacy of the multi-view system over the single-view one. For glasses, it provides a substantial improvement in accuracy by about 50\% and 40\% for the generic scenario and all scenarios, respectively. In addition, it brings $\sim$17\% enhancement in estimation availability. For contact lenses, single-camera system standalone yields a high accuracy and availability. Yet, a multi-view system further enhances the accuracy and availability. 

\begin{table} [t!]
\vspace{-3mm}
\caption{Performance comparison on the same subject.}
\vspace{-2mm}
\label{contact_vs_glasses_murat}
\centering
\begin{tabular}{l|cccc|cccc}
\hline\noalign{\smallskip}
\multirow{3}{*}{\parbox{0.2cm}{Eyewear}} & \multicolumn{4}{c|}{Single-cam view} & \multicolumn{4}{c}{Multi-view}\\ 
& \multicolumn{2}{c}{Exp 2} & \multicolumn{2}{c|}{Exp [0-7]} & \multicolumn{2}{c}{Exp 2} & \multicolumn{2}{c}{Exp [0-7]} \\
 	& ($^\circ$) & 	(\%) & ($^\circ$) & 	(\%) & ($^\circ$) & 	(\%) & ($^\circ$) & 	(\%)\\ 
\noalign{\smallskip}\hline\noalign{\smallskip}
Contacts & 0.99 & 96.1 & 1.18 & 95.1 & 0.76 & 100 & 0.97 & 99.9\\ 
Glasses    & 2.1  & 84 & 2.31 & 82 & 1.08 & 100 & 1.53 & 99  \\ 
\noalign{\smallskip}\hline
\end{tabular}
\vspace{-6mm}
\end{table}

\begin{figure}[b!]
\centering
\subcaptionbox{\label{fig_asian_dark}}{\includegraphics[width=0.45\columnwidth]{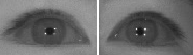}}
\hfill
\subcaptionbox{\label{fig_asian_dark_gl}}{\includegraphics[width=0.45\columnwidth]{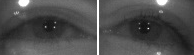} }
\hfill
\subcaptionbox{\label{fig_caucasian_dark}}{\includegraphics[width=0.45\columnwidth]{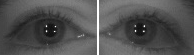}}
\hfill
\subcaptionbox{\label{fig_caucasian_light_gl}}{\includegraphics[width=0.45\columnwidth]{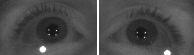}}
\caption{Sample eye appearances: (a) Asian dark eyes without glasses, (b) Asian dark eyes with glasses, (c) Caucasian dark eyes without glasses, (d) Caucasian dark eyes with glasses.}
\label{fig_asian_caucasian}
\end{figure}

\subsubsection{Eye Type Robustness}
\label{eye_type_robustness}
Variations in eye type, e.g., iris color, eye shape, pupil response, may significantly affect the performance of eye trackers \cite{Nguyen2002}. As our dataset contains eye type variations (Fig.~\ref{fig_asian_caucasian}), we evaluate the proposed framework's robustness to iris color and eye shape. Firstly, since the iris color has a great influence on both the pupil size and opening of eyelids when exposed to various illumination conditions, we categorized the subjects into two groups according to the iris color such as dark-eyes (10 subjects) and light-eyes (10 subjects). On average over all experiments, dark- and light-eyes groups achieve 1.15$^\circ$ with 96.2$\%$ availability and 1.44$^\circ$ with 93.3$\%$ performances, respectively. However, the results may be biased towards the dark-eyes group since most of its subjects do not wear glasses. An interesting result is that the performance difference between the light-eyes (1.33$^\circ$) and dark-eyes groups (0.92$^\circ$) is especially large under sunlight (experiment $\#$0). The reason is that the pupil size and eye opening are more significantly affected for light-eyes in comparison with the dark-eyes due to their higher sensitivity to the sunlight. Table~\textcolor{red}{S-II} in the supplementary information shows average estimation accuracy and estimation availability in detail for each experiment in various categories. 

It is also important to note that the pupil detection method has an influence on the robustness to eye color variations. As mentioned earlier and discussed in detail in \cite{ArarThesis2017}, bright-pupil based method is frequently employed by the previous work as the feature detection is simpler compared to dark-pupil based one. However, in our preliminary experiments, we observed that dark-pupil based feature detection is less sensitive to the variations in eye color. In bright-pupil based method, the accuracy of the pupil detection heavily relies on the pupil response (brightness), which is highly affected by users' momentary pupil size that varies according to the eye color, ethnicity, and ambient illumination. Therefore, in the final framework, we employ dark-pupil based feature detection for becoming less sensitive to eye type and illumination factors. In our feature work, we aim to give a more structured quantitative comparison on pupil detection methods. 
 
Furthermore, we categorized the subjects by their eye shape into two groups: Asian eyes (2 subjects) and non-Asian eyes (18 subjects) to analyze the impact of the eye shape. Our results show that Asian eyes (1.58$^\circ$ with 93.35$\%$ availability) perform worse than non-Asian eyes (1.25$^\circ$ with 97.7$\%$ availability). Yet, the system can still accurately estimate the gaze for our Asian subjects. The decrease in the availability may indicate that the feature detection for them might be more challenging due to the eye shape. Nevertheless, it is difficult to make a strong conclusion as the two sets are imbalanced. In addition, we note that there is a significant variation across Asian eyes including round, narrow, almond, hooded, triangular, prominent, or deep-set eye shapes \cite{Kiranantawat2015}. In addition, the eyes can be a single eyelid, low/incomplete eyelid crease, and double eyelid. For some of these eye shapes (e.g., narrow, hooded), eye tracking could be highly challenging as the creation and detection of the gaze features could be exigent. Our dataset currently do not contain sufficient eye type variation. Hence, we plan to recruit a larger number of Asian subjects and increase the variation in eye type to obtain a more reliable analysis.

\setcounter{footnote}{0}

\begin{table*}[t!]
\centering
\label{my-label}
\vspace{-3mm}
\resizebox*{!}{0.4\linewidth}{
\begin{tabular}{p{0.05cm}l|ccc|ccc|cccc|c}
\multicolumn{2}{c|}{\multirow{2}{*}{{\bf Method}}} & \multicolumn{3}{c|}{{\bf Hardware Setup}} & \multicolumn{3}{c|}{{\bf Accuracy}} &  \multicolumn{4}{c|}{{\bf Robustness}} & {\bf FoV} \\
\multicolumn{2}{c|}{} & \multicolumn{1}{c}{Cam(s)} & \multicolumn{1}{c}{Light(s)} & \multicolumn{1}{c|}{Calib.} & SH($^\circ$) & MH($^\circ$) & Eye Data & HP & Head Movement & Eyewear & Illum. & FL\\ \hline

\multirow{5}{*}{\rotatebox[origin=c]{90}{\bf{Appearance}}}
 & Zhang et al (2015) \cite{Zhang2015} & 1 & - & - & 6.3\footnotemark[1] & ? & Both & Free & ?  & \cmark  & \cmark & ?  \\
 & Wood et al (2016) \cite{Wood2016a} & 1 & - & - & 9.95\footnotemark[2] & ? & ? & Free & ? & \cmark & \cmark & ?  \\ 
 & Lu et al (2015) \cite{Lu2015} & 1 & - & - & 2.5 & 9.65 & Single & Free & (40 x ? x ? ) & - & - & ? \\
 & Mora \& Odobez (2016) \cite{FunesMora2016} & 1+Kinect & 5 & pre & 1.9\footnotemark[3] & 3.5 & Both & Free & ? & - & - & 6.1 \\
 & Krafka et al (2016) \cite{Krafka2016} & 1 & - & - & $\sim$3.5\footnotemark[4] & ? & Both & Free & ? & \cmark & \cmark & ?  \\ \hline

\multirow{6}{*}{\rotatebox[origin=c]{90}{\bf{3D Model}}}
 & Beymer \& Flickner (2003) \cite{Beymer2003} & 4* & 2 & fully & $\sim$0.6 & - & Single & Free & Limited & - & - & 4.8\\
 & Hennessey et al (2006)\cite{Hennessey2006} & 1+1 & 3 & fully & $\sim$1 & $<$1 & Single & Free & (14 x 12 x 20) & - & - & 32 \\
 & Guestrin \& Eizenman (2007) \cite{Guestrin2007} & 2 & 4 & fully & -  & $\sim$1 & Single & Free & (10 x 8 x 10) & - & - & 35 \\
 & Park (2007) \cite{Park2007} & 1+2* & 4 & fully & -  & $\sim$1 & Both & Free & (? x ? x 20) & \cmark & - & ? \\
 & Lai et al (2015) \cite{Lai2015} & 2 & 2 & fully & -  & $\sim$1 & Single & Free & (10 x 5 x 10) & - & - & 37  \\
 & Sun et al (2015) \cite{Sun2015} & Kinect & ? & pre & $\sim$1.5  & $\sim$2 & Single & Free & (20 x 20 x 8) & - & - & 6.1 \\\hline
 
\multirow{4}{*}{\rotatebox[origin=c]{90}{\bf{Regression}}}
 &  Zhu \& Ji (2007) \cite{Zhu2007} & 2 & 2 & fully & $\sim$1.1 & $\sim$1.8 & Single & Free & (20 x 20 x 30) & - & - & ? \\
 & Cerrolaza et al (2008) \cite{Cerrolaza2008} & 1 & 2 & - & $\sim$1 & $\sim$1 & Single & Fixed & (0 x 0 x 10) & - & - & 35  \\
 & Sesma-sanchez et al (2012) \cite{Sesma-sanchez2012} & 1 & 2 & - & $\sim$1 & $\sim$1 & Both & Fixed & (0 x 0 x 6) & - & - & 35  \\
 & Cerrolaza et al (2012) \cite{Cerrolaza2012} & 1 & 2 & - & $\sim$0.9 & $\sim$1.3 & Single & Fixed & (0 x 0 x 12) & - & - & 16  \\\hline
 
\multirow{7}{*}{\rotatebox[origin=c]{90}{\bf{Cross ratio}}}
 & Yoo \& Chung (2005) \cite{Yoo2005} & 1+1* & 4+1 & - & $\sim$1.6 & - & Single & Free & Limited & - & - & ? \\
 & Hansen et al (2010) \cite{Hansen2010} & 1 & 4 & - & $\sim$1 & - & Single & Free & Limited & - & - & ?  \\
 & Coutinho \& Morimoto (2013) \cite{Coutinho2013} & 1 & 4+1 & - & $\sim$0.4 & $\sim$0.5 & Single & Fixed & (25 x ? x 25) & - & - & $>$ 35 \\
 & Zhang \& Cai (2014) \cite{Zhang2014} & 1 & 8 & - & $\sim$0.4 & $\sim$0.6 & Both & Fixed & (10 x ? x 20) & - & - & 13 \\
 & Huang et al (2014) \cite{Huang2014} & 1 & 8 & - & $\sim$0.8 & $\sim$1.6 & Single & Fixed & (? x ? x 20) & - & - & 13 \\
 & Arar et al (2015) \cite{Arar2015f} (multi-view)& 3 & 4+1 & - & 0.86 & - & Both & Free & Limited & - & - & 12\footnotemark[5] \\
 & {\bf Proposed multi-view framework} & 3 & 4 & - & {\bf 0.99} & {\bf 1.27} & {\bf Both} & {\bf Free} & {\bf (30 x ? x 20)} & {\bf \cmark} & {\bf \cmark} & {\bf 8\footnotemark[5]} \\\hline
  
\end{tabular}

}
\vspace{-1mm}
\caption{Comparison of existing eye tracking systems. In \emph{"Cam(s)"} column, $^*$ indicates that a pan-tilt unit is employed. \emph{"Calib."} column indicates whether explicit camera and scene geometry calibrations are required: "fully" means both are required, "pre" means the sensor is pre-calibrated. In \emph{"Accuracy"}, \emph{"SH"} and \emph{"MH"} correspond to stable and moving head scenarios, respectively. The results refer to, unless stated otherwise, person-specific scenarios on within-dataset evaluations. \emph{"HP"} column indicates whether users' head pose were fixed, e.g., using a chinrest. In \emph{"FoV"} column, the systems' working volume is presented by \emph{"FL"}, focal length in mm. The smaller the focal length, the larger the FoV.}
\label{comparison_with_previous_work}
\vspace{-5mm}
\end{table*}

\vspace{-4mm}
\subsection{Real-time Implementation}
\label{cge:s:real_time_tracking}
The proposed gaze estimation system and methods are developed in C++\footnote{\url{https://lts5www.epfl.ch/eyetracking}}. OpenCV library is mainly used for image processing and computer vision algorithms. Localization of facial landmarks were performed using a supervised descent method (SDM)-based face tracker. Furthermore, to achieve real-time tracking performance, OpenMP application programming interface was utilized for the parallellization of our library implementation. The computational complexity of the system is lower than \emph{3D model-based} methods as the gaze estimation relies on perspective geometry transformations and computation of cross ratios. This enables to achieve a real-time implementation without requiring any particular performance optimization. In our implementation, the most computationally expensive process is face tracking. Cross ratio-based gaze estimation on both eyes, user calibration, and adaptive fusion processes require much lower computational effort. For instance, these three processes take only $\sim$8 ms on a PC with Intel i7 3.2GHz processor, whereas face tracking solely takes $\sim$24 ms. Our current three-camera prototype system can simultaneously output \ac{PoR}s for both eyes as well as an overall \ac{PoR} at $\sim$30 fps with a mean estimation accuracy error of $\sim$1$^\circ$ of visual angle. Yet, we note that there is much room for computationally improving our implementation to reach higher frame rates. As future work, we plan to replace the current computationally expensive face tracker with a faster one, such as local binary features (LBF)-based face tracking \cite{Ren2014}, one millisecond face alignment with an ensemble of regression trees \cite{Kazemi2014}. In addition, we plan to further optimize feature detection and adaptive fusion implementations.

\vspace{-3mm}
\section{Discussion}
\label{discussion}
Future directions in eye tracking research, towards becoming a pervasive technology, should not only focus on achieving high estimation accuracies, but also on having robustness against real-world settings such as natural head pose changes, large head movements, varying illumination conditions, use of eyewear, and between-subject eye type variations. Besides, having a convenient user calibration, flexible hardware setup, minimal setup calibration, low complexity and cost should be taken into consideration as important evaluation criteria. In this regard, in Section~\ref{related_work}, we describe various eye tracking techniques, analyze their pros and cons with respect to each other, and discuss whether they satisfy some of the aforementioned criteria. Therefore, the best, in other words, the most appropriate, approach depends on the application requirements. In this work, we mainly target eye tracking scenarios that require high-accuracy ($\sim$1$^\circ$) estimation and robustness, e.g., gaze-based controlling, typing and navigation. To achieve our accuracy and robustness goals, we design a novel multi-camera framework, which tracks users' gaze simultaneously from various views, and then combines the acquired gaze information from all sensors using an adaptive fusion mechanism to output an overall \ac{PoR}. In comparison with conventional single-view approaches, multi-view tracking enables a more reliable gaze features detection even under challenging scenarios. In addition, owing to the proposed adaptive fusion mechanisms, the framework achieves high accuracies and robustnmess under real-world conditions.

A comparison of previous work in various aspects such as hardware setup and calibration requirements, accuracy, robustness, and working volume, is given in Table~\ref{comparison_with_previous_work}. Since the majority of existing efforts requires particular hardware and system setups, e.g., additional light sources, setup calibration, use of 3D or depth information, we could not reproduce and validate the reported performances for all. Instead, for these, we reported the performances directly from the corresponding references. Although a direct numerical comparison would not be completely fair, the provided information can still help us to make the following inferences. First of all, we observe that the popularity of \emph{appearance-based} methods, which have lower hardware and calibration requirements, have been increasing recently in parallel with the recent advancements in machine learning, e.g., convolutional neural networks (CNNs), and in the synthesizing and rendering technology. Despite the fact that their accuracies and head movement tolerances are still not comparable to those of \emph{feature-based} methods, their potential is likely to be exploited in the foreseeable future. 

On the other hand, even though \emph{feature-based} methods outperform \emph{appearance-based}, the bothersome hardware requirements, e.g., \ac{NIR} cameras and light sources, remain an important concern. The setup complexity is especially high for \emph{3D model-based} systems, such that fully-calibrated setups consisting of multiple cameras or a Kinect-like sensor are required for accurate 3D modelling. \emph{Cross ratio-based} systems and most of \emph{regression-based} systems have an important advantage over \emph{3D model-based} ones. They require uncalibrated setups and less complex (2D) eye models while providing competitive accuracies to \emph{3D model-based} systems. Among these, it is also clear that there is an accuracy gap between fixed-head (using chinrest) and free-head eye tracking since the approximated eye models are sensitive to head movements.


Moreover, the results indicate that tracking performances significantly benefit from high-resolution eye data. For example, Coutinho \& Morimoto \cite{Coutinho2013} reported an impressive accuracy, about 0.5$^\circ$, under large head movements through planarization of gaze features. Nonetheless, their system required eye resolution of 640$\times$480 pixels, that is 7-fold of ours. They captured eye data using a narrow \ac{FoV} lens and used a chinrest to keep users' eye within the \ac{FoV} of the camera. In addition, Huang et al. \cite{Huang2014} and Zhang \& Cai \cite{Zhang2014} proposed two alternative methods that are highly effective to compensate for the head movements, while requiring relatively lower resolution eye data, i.e., 13-mm focal lenses were used. However, similar to \cite{Coutinho2013}, they both utilized a chinrest during their evaluation. Since use of chinrest is unnatural for users and represents an unrealistic tracking scenario, it remains an important limitation of their evaluations. On the contrary, our methodology allows for not only head translations but also head rotations while requiring lower resolution eye data ($\sim$90$\times$50 pixels) captured using 8-mm lenses. Lower resolution data naturally results in a lower accuracy, yet, the proposed adaptive fusion mechanism successfully closes the accuracy gap by effectively combining the gaze outputs obtained by multiple sensors. Besides, our system accounts for eyewear and illumination robustness, which have largely been neglected by the majority of the previous efforts.

As depicted from Table~\ref{comparison_with_previous_work}, the proposed multi-view approach and implemented prototype system achieves a competitive accuracy while offering more robustness to aforementioned real-world conditions. Still, further improvements on accuracy, robustness or setup complexity can be achieved through certain hardware and software modifications. For instance, explicit head movement compensation techniques, such as learning an adaptive homography from simulated data \cite{Huang2014} or planarization of features \cite{Coutinho2013}, can be employed to further improve the head movement robustness. In addition, as utilized by most commercial eye trackers, certain hardware solutions, e.g., auto-focus lenses or smart dynamic illumination techniques, can greatly enhances the estimation accuracy and availability.

\footnotetext[1]{Person-independent within-dataset evaluation on MPIIGaze dataset \cite{Zhang2015}.}
\footnotetext[2]{Person-independent cross-dataset evaluation on MPIIGaze dataset \cite{Zhang2015}.}
\footnotetext[3]{Person-specific within-dataset evaluation on Eyediap dataset \cite{FunesMora2014}.}
\footnotetext[4]{Person-specific within-dataset evaluation on GazeCapture dataset \cite{Krafka2016}.}
\footnotetext[5]{Single camera property. Multi-view setup has a significantly larger FoV.}


As the proposed multi-view framework is independent of the gaze estimation method used, alternative gaze estimation methods can also be integrated with regard to the application requirements. Although \emph{cross ratio-based} method is suggested for high-accuracy estimation on desktop scenarios due to its particular advantages, such as enabling high-accuracy using an uncalibrated and flexible setup, an \emph{appearance (CNN)-based} method (e.g., Zhang et al. \cite{Zhang2015}, Krafka et al. \cite{Krafka2016}) can be used to lower the setup complexity, or a \emph{3D model-based} method can be implemented to achieve even higher accuracies. Furthermore, the number of cameras and their configuration can be varied according to the application scenario without requiring further system adjustments, particularly for uncalibrated settings. For instance, the current prototype can easily be configured to work under challenging tracking scenarios, such as in-car driving scenarios, children's eye tracking, or customized eye trackers for disabled people.


\vspace{-3mm}
\section{Conclusions}
\label{conclusion}
This paper presents a novel multi-view eye tracking framework to revisit the robustness concerns in eye tracking, particularly to head movements and glasses. Instead of computing the user gaze from a single view as performed by the previous work, leveraging multiple eye appearances simultaneously acquired from various views provides with enhanced estimation accuracy and robustness under challenging real-world conditions. The main advantage of the multi-view approach is that for each frame, we calculate multiple gaze outputs using the features extracted from various eye appearances. This enables to extract the features more reliably even under challenging conditions, where they are obstructed in the conventional single-camera view. Under large head movements and use of glasses, our evaluations show that the proposed approach improves the tracking performance of a single-camera setup by about 20\% (0.2-0.6$^\circ$) in estimation accuracy and 10-20\% in estimation availability. The results also demonstrate that our approach is highly tolerant to illumination and between-subject eye type variations. In addition to the improved robustness to challenging conditions, the system's overall accuracy greatly benefits from the multi-view setup under normal conditions. The proposed methodology provides by about 30\% improvement in accuracy, owing to the adaptive fusion mechanisms, which account for the reliability of estimations determined from user's predicted gazing behavior and momentary head poses with respect to each camera. The current implementation runs at 30 fps, obtains $\sim$1$^\circ$ estimation accuracy error and nearly 100\% estimation availability under challenging experimental scenarios, which makes it appropriate for high-accuracy demanding applications.



%

%

\vspace{-3mm}

\ifCLASSOPTIONcaptionsoff
  \newpage
\fi

\bibliographystyle{IEEEtran}
\bibliography{IEEEabrv,main}

\begin{thebibliography}{10}
\providecommand{\url}[1]{#1}
\csname url@samestyle\endcsname
\providecommand{\newblock}{\relax}
\providecommand{\bibinfo}[2]{#2}
\providecommand{\BIBentrySTDinterwordspacing}{\spaceskip=0pt\relax}
\providecommand{\BIBentryALTinterwordstretchfactor}{4}
\providecommand{\BIBentryALTinterwordspacing}{\spaceskip=\fontdimen2\font plus
\BIBentryALTinterwordstretchfactor\fontdimen3\font minus
  \fontdimen4\font\relax}
\providecommand{\BIBforeignlanguage}[2]{{%
\expandafter\ifx\csname l@#1\endcsname\relax
\typeout{** WARNING: IEEEtran.bst: No hyphenation pattern has been}%
\typeout{** loaded for the language `#1'. Using the pattern for}%
\typeout{** the default language instead.}%
\else
\language=\csname l@#1\endcsname
\fi
#2}}
\providecommand{\BIBdecl}{\relax}
\BIBdecl

\bibitem{Underwood2005}
G.~Underwood, \emph{Cognitive processes in eye guidance}.\hskip 1em plus 0.5em
  minus 0.4em\relax Oxford Pr., 2005.

\bibitem{Duchowski2007}
A.~T. Duchowski, \emph{{Eye tracking methodology}}.\hskip 1em plus 0.5em minus
  0.4em\relax Springer, 2007.

\bibitem{Survey_Hansen2010}
D.~W. Hansen and Q.~Ji, ``{In the eye of the beholder: a survey of models for
  eyes and gaze},'' \emph{PAMI}, vol.~32, no.~3, pp. 478--500, 2010.

\bibitem{Beymer2003}
\BIBentryALTinterwordspacing
D.~Beymer and M.~Flickner, ``{Eye gaze tracking using an active stereo head},''
  in \emph{CVPR}, 2003, pp. 451--458.
\BIBentrySTDinterwordspacing

\bibitem{Hennessey2006}
\BIBentryALTinterwordspacing
C.~Hennessey, B.~Noureddin, and P.~Lawrence, ``{A single camera eye-gaze
  tracking system with free head motion},'' \emph{Measurement}, 2006.
\BIBentrySTDinterwordspacing

\bibitem{Guestrin2007}
\BIBentryALTinterwordspacing
E.~D. Guestrin and M.~Eizenman, ``{Remote Point-of-Gaze Estimation with Free
  Head Movements Requiring a Single-Point Calibration},'' in \emph{Proc. Int.
  Conf. Eng. Med. Biol. Soc.}, 2007, pp. 4556--4560.
\BIBentrySTDinterwordspacing

\bibitem{Park2007}
K.~R. Park, ``{A real-time gaze position estimation method based on a 3-D eye
  model.}'' \emph{TSMC Part B}, vol.~37, no.~1, pp. 199--212, 2007.

\bibitem{Sun2015}
\BIBentryALTinterwordspacing
L.~Sun, Z.~Liu, and M.~T. Sun, ``{Real time gaze estimation with a consumer
  depth camera},'' \emph{Inf. Sci. (Ny).}, vol. 320, pp. 346--360, 2015.
\BIBentrySTDinterwordspacing

\bibitem{Zhu2006}
Z.~Zhu, Q.~Ji, and K.~P. Bennett, ``{Nonlinear eye gaze mapping function
  estimation via support vector regression},'' in \emph{ICPR}, 2006, pp.
  1132--1135.

\bibitem{Zhu2007}
\BIBentryALTinterwordspacing
Z.~Zhu and Q.~Ji, ``{Novel Eye Gaze Tracking Techniques Under Natural Head
  Movement},'' \emph{TBE}, vol.~54, no.~12, 2007.
\BIBentrySTDinterwordspacing

\bibitem{Cerrolaza2008}
J.~J. Cerrolaza, A.~Villanueva, and R.~Cabeza, ``{Taxonomic study of polynomial
  regressions applied to the calibration of video-oculographic systems},'' in
  \emph{ETRA}, 2008.

\bibitem{Sesma-sanchez2012}
L.~Sesma-sanchez, A.~Villanueva, and R.~Cabeza, ``{Gaze Estimation
  Interpolation Methods Based on Binocular Data},'' \emph{TBE}, vol.~59, no.~8,
  pp. 2235--2243, 2012.

\bibitem{Yoo2005}
\BIBentryALTinterwordspacing
D.~H. Yoo and M.~J. Chung, ``{A novel non-intrusive eye gaze estimation using
  cross-ratio under large head motion},'' \emph{CVIU}, vol.~98, no.~1, 2005.
\BIBentrySTDinterwordspacing

\bibitem{Hansen2010}
\BIBentryALTinterwordspacing
D.~Hansen, J.~Agustin, and A.~Villanueva, ``{Homography normalization for
  robust gaze estimation in uncalibrated setups},'' in \emph{ETRA}, 2010.
\BIBentrySTDinterwordspacing

\bibitem{Coutinho2013}
\BIBentryALTinterwordspacing
F.~L. Coutinho and C.~Morimoto, ``{Improving head movement tolerance of
  cross-ratio based eye trackers},'' \emph{IJCV}, vol. 101, no.~3, 2013.
\BIBentrySTDinterwordspacing

\bibitem{Zhang2014}
\BIBentryALTinterwordspacing
Z.~Zhang and Q.~Cai, ``{Improving cross-ratio-based eye tracking techniques by
  leveraging binocular fixation constraint},'' in \emph{ETRA}, 2014.
\BIBentrySTDinterwordspacing

\bibitem{Huang2014}
\BIBentryALTinterwordspacing
J.-B. Huang, Q.~Cai, Z.~Liu, N.~Ahuja, and Z.~Zhang, ``{Towards accurate and
  robust cross-ratio based gaze trackers through learning from simulation},''
  in \emph{ETRA}, 2014, pp. 75--82.
\BIBentrySTDinterwordspacing

\bibitem{Arar2016j}
N.~M. Arar, H.~Gao, and J.~P. Thiran, ``A regression-based user calibration
  framework for real-time gaze estimation,'' \emph{TCSVT}, vol.~27, no.~12, pp.
  2623--2638, 2017.

\bibitem{Arar2015f}
------, ``Robust gaze estimation based on adaptive fusion of multiple
  cameras,'' in \emph{IEEE Int. Conf. Aut. Face \& Gesture Recogn. (FG)}, 2015.

\bibitem{Kang2007}
J.~J. Kang, E.~D. Guestrin, W.~J. Maclean, and M.~Eizenman, ``{Simplifying the
  cross-ratios method of point-of-gaze estimation},'' in \emph{Can. Med. Biol.
  Eng. Conf.}, 2007.

\bibitem{Villanueva2008}
A.~Villanueva and R.~Cabeza, ``{A novel gaze estimation system with one
  calibration point},'' \emph{TSMC Part B}, vol.~38, no.~4, pp. 1123--1138,
  2008.

\bibitem{Arar2015w}
N.~M. Arar, H.~Gao, and J.~P. Thiran, ``Towards convenient calibration for
  cross-ratio based gaze estimation,'' in \emph{WACV}, 2015, pp. 642--648.

\bibitem{Lai2015}
C.~C. Lai, S.~W. Shih, and Y.~P. Hung, ``{Hybrid method for 3-D gaze tracking
  using glint and contour features},'' \emph{TCSVT}, vol.~25, no.~1, 2015.

\bibitem{Cerrolaza2012}
\BIBentryALTinterwordspacing
J.~Cerrolaza, A.~Villanueva, M.~Villanueva, and R.~Cabeza, ``{Error
  characterization compensation in eye tracking systems},'' in \emph{ETRA},
  2012.
\BIBentrySTDinterwordspacing

\bibitem{Lu2015}
\BIBentryALTinterwordspacing
F.~Lu, Y.~Sugano, T.~Okabe, and Y.~Sato, ``{Gaze Estimation From Eye
  Appearance: A Head Pose-Free Method via Eye Image Synthesis},'' \emph{TIP},
  vol.~24, no.~11, pp. 3680--3693, 2015.
\BIBentrySTDinterwordspacing

\bibitem{Zhang2015}
\BIBentryALTinterwordspacing
X.~Zhang, Y.~Sugano, M.~Fritz, and A.~Bulling, ``{Appearance-based gaze
  estimation in the wild},'' in \emph{CVPR}, 2015, pp. 4511--4520.
\BIBentrySTDinterwordspacing

\bibitem{Krafka2016}
\BIBentryALTinterwordspacing
K.~Krafka, A.~Khosla, P.~Kellnhofer, H.~Kannan, S.~Bhandarkar, Matusik, and
  A.~Torralba, ``{Eye tracking for everyone},'' in \emph{CVPR}, 2016.
\BIBentrySTDinterwordspacing

\bibitem{Wood2016a}
\BIBentryALTinterwordspacing
E.~Wood, T.~Baltru{\v{s}}aitis, L.-P. Morency, P.~Robinson, and A.~Bulling,
  ``{Learning an appearance-based gaze estimator from one million synthesised
  images},'' in \emph{ETRA}, 2016, pp. 131--138.
\BIBentrySTDinterwordspacing

\bibitem{Guestrin2006}
\BIBentryALTinterwordspacing
E.~Guestrin and M.~Eizenman, ``{General Theory of Remote Gaze Estimation Using
  the Pupil Center and Corneal Reflections},'' \emph{TBE}, vol.~53, no.~6, pp.
  1124--1133, jun 2006.
\BIBentrySTDinterwordspacing

\bibitem{Utsumi2012}
\BIBentryALTinterwordspacing
A.~Utsumi, K.~Okamoto, N.~Hagita, and K.~Takahashi, ``{Gaze tracking in wide
  area using multiple camera observations},'' in \emph{ETRA}, 2012.
\BIBentrySTDinterwordspacing

\bibitem{Nagamatsu2011}
T.~Nagamatsu, R.~Sugano, Y.~Iwamoto, J.~Kamahara, and N.~Tanaka,
  ``{User-calibration-free gaze estimation method using a binocular 3D eye
  model},'' \emph{IEICE Trans. Inf. Syst.}, vol. E94-D, no.~9, pp. 1817--1829,
  2011.

\bibitem{Sun2014}
\BIBentryALTinterwordspacing
L.~Sun, M.~Song, Z.~Liu, and M.~Sun, ``{Real-time gaze estimation with online
  calibration},'' \emph{IEEE Multimedia}, vol.~21, no.~4, pp. 28--37, 2014.
\BIBentrySTDinterwordspacing

\bibitem{Chen2015}
J.~Chen and Q.~Ji, ``{A probabilistic approach to online eye gaze tracking
  without explicit personal calibration},'' \emph{TIP}, vol.~24, no.~3, 2015.

\bibitem{White1993}
K.~White, T.~Hutchhinson, and J.~Carley, ``{Spatially Dynamic Calibration of an
  Eye-Tracking System},'' \emph{TSMC}, vol.~23, no.~4, pp. 1162--1168, 1993.

\bibitem{Ebisawa1998}
\BIBentryALTinterwordspacing
Y.~Ebisawa, ``{Improved video-based eye-gaze detection method},'' \emph{IEEE
  Trans. Instrum. Meas.}, vol.~47, no.~4, pp. 948--955, 1998.
\BIBentrySTDinterwordspacing

\bibitem{Ji2002}
\BIBentryALTinterwordspacing
Q.~Ji and X.~Yang, ``{Real-time eye, gaze, and face pose tracking for
  monitoring driver vigilance},'' \emph{Real-Time Imaging}, vol.~8, no.~5,
  2002.
\BIBentrySTDinterwordspacing

\bibitem{Villanueva2007}
\BIBentryALTinterwordspacing
A.~Villanueva and R.~Cabeza, ``{Models for gaze tracking systems},''
  \emph{Eurasip J. Image Video Process.}, vol. 2007, pp. 1--16, 2007.
\BIBentrySTDinterwordspacing

\bibitem{Kubler2016}
\BIBentryALTinterwordspacing
T.~C. K{\"{u}}bler, T.~Rittig, E.~Kasneci, J.~Ungewiss, and C.~Krauss,
  ``{Rendering refraction and reflection of eyeglasses for synthetic eye
  tracker images},'' in \emph{ETRA}, 2016, pp. 143--146.
\BIBentrySTDinterwordspacing

\bibitem{sdm}
\BIBentryALTinterwordspacing
X.~Xiong and F.~{De la Torre}, ``{Supervised Descent Method and Its
  Applications to Face Alignment},'' in \emph{CVPR}, 2013, pp. 532--539.
\BIBentrySTDinterwordspacing

\bibitem{ArarThesis2017}
N.~M. Arar, ``Robust eye tracking based on adaptive fusion of multiple
  cameras,'' Ph.D. dissertation, EPFL, 2017, th{\`{e}}se no 7933.

\bibitem{Zangemeister1982}
W.~H. Zangemeister and L.~Stark, ``Types of gaze movement: variable
  interactions of eye and head movements.'' \emph{Experimental neurology},
  vol.~77, no.~3, pp. 563--77, 1982.

\bibitem{Saragih2009}
J.~M. Saragih, S.~Lucey, and J.~F. Cohn, ``Face alignment through subspace
  constrained mean-shifts,'' in \emph{ICCV}, 2009, pp. 1034--1041.

\bibitem{Martin2008}
\BIBentryALTinterwordspacing
M.~B{\"{o}}hme, M.~Dorr, M.~Graw, T.~Martinetz, and E.~Barth, ``{A software
  framework for simulating eye trackers},'' in \emph{ETRA}, 2008, p. 251.
\BIBentrySTDinterwordspacing

\bibitem{Schaeffel2006}
\BIBentryALTinterwordspacing
F.~Schaeffel, ``{Myopia: The Importance of Seeing Fine Detail},'' \emph{Curr.
  Biol.}, vol.~16, no.~7, pp. 257--259, apr 2006.
\BIBentrySTDinterwordspacing

\bibitem{Nguyen2002}
\BIBentryALTinterwordspacing
K.~Nguyen, C.~Wagner, D.~Koons, and M.~Flickner, ``{Differences in the infrared
  bright pupil response of human eyes},'' in \emph{ETRA}, 2002, p. 133.
\BIBentrySTDinterwordspacing

\bibitem{Kiranantawat2015}
K.~Kiranantawat, J.~Suhk, and A.~Nguyen, ``The asian eyelid: Relevant
  anatomy,'' \emph{Seminars in Plastic Surgery}, vol.~29, no.~3, 2015.

\bibitem{FunesMora2016}
\BIBentryALTinterwordspacing
K.~A. Mora and J.-M. Odobez, ``{Gaze Estimation in the 3D Space Using RGB-D
  Sensors},'' \emph{IJCV}, vol. 118, no.~2, pp. 194--216, jun 2016.
\BIBentrySTDinterwordspacing

\bibitem{Ren2014}
\BIBentryALTinterwordspacing
S.~Ren, X.~Cao, Y.~Wei, and J.~Sun, ``{Face Alignment at 3000 FPS via
  Regressing Local Binary Features},'' in \emph{CVPR}, 2014, pp. 1685--1692.
\BIBentrySTDinterwordspacing

\bibitem{Kazemi2014}
\BIBentryALTinterwordspacing
V.~Kazemi and J.~Sullivan, ``{One millisecond face alignment with an ensemble
  of regression trees},'' in \emph{CVPR}, 2014, pp. 1867--1874.
\BIBentrySTDinterwordspacing

\bibitem{FunesMora2014}
\BIBentryALTinterwordspacing
K.~A. {Mora} and J.-M. Odobez, ``{Geometric generative gaze estimation (G3E)
  for remote RGB-D cameras},'' in \emph{CVPR}, 2014, pp. 1773--1780.
\BIBentrySTDinterwordspacing

\end{thebibliography}

\includepdf[pages={1-}]{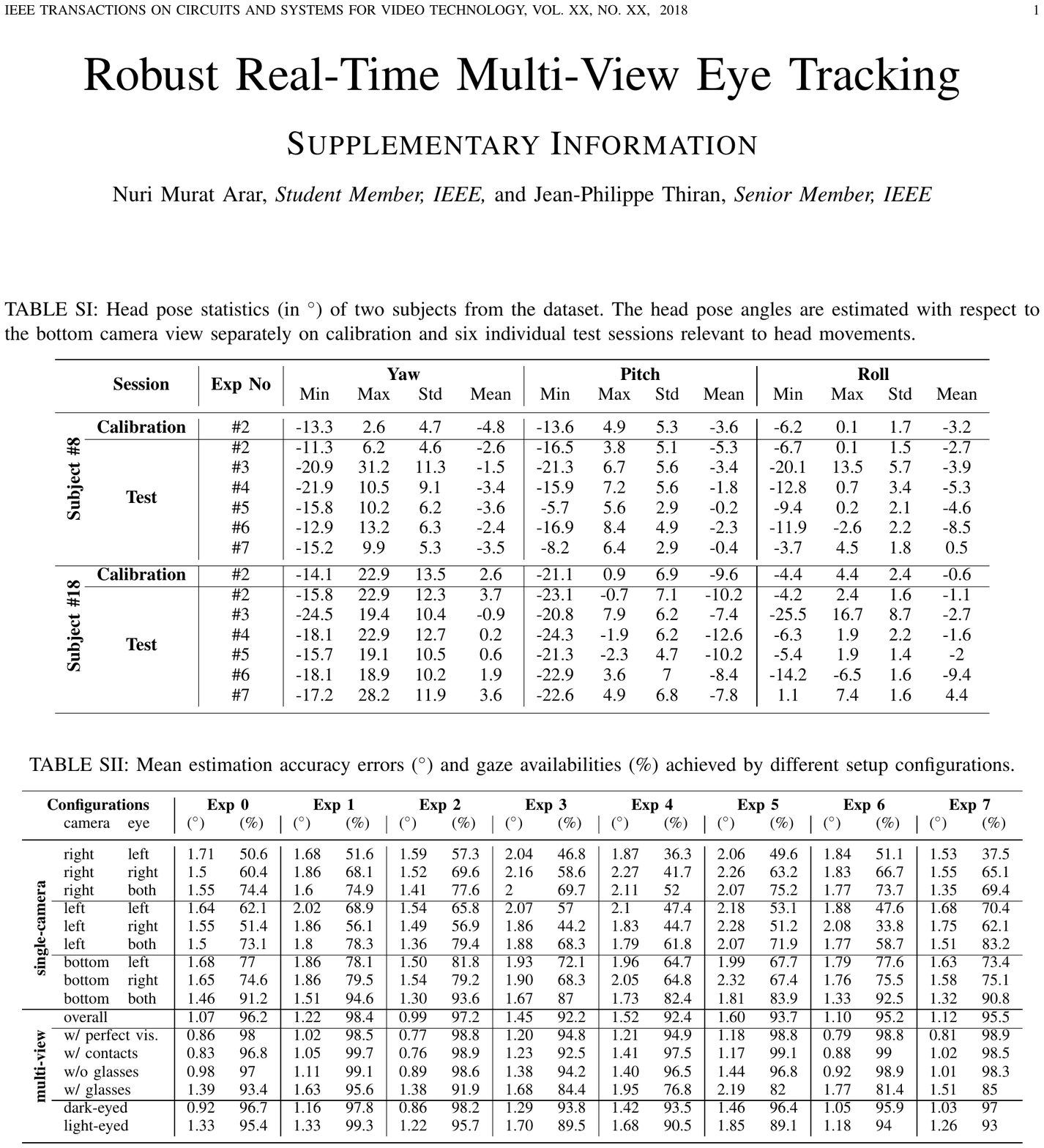}

\end{document}